# Making a Spiking Net Work: Robust brain-like unsupervised machine learning

Peter G. Stratton, Andrew Wabnitz, Chip Essam, Allen Cheung, and Tara J. Hamilton, *Member, IEEE*

*Abstract*— **The surge in interest in Artificial Intelligence (AI) over the past decade has been driven almost exclusively by advances in Artificial Neural Networks (ANNs). While ANNs set state-of-the-art performance for many previously intractable problems, the use of global gradient descent necessitates large datasets and computational resources for training, potentially limiting their scalability for real-world domains. Spiking Neural Networks (SNNs) are an alternative to ANNs that use more brain-like artificial neurons and can use local unsupervised learning to rapidly discover sparse recognizable features in the input data. SNNs, however, struggle with dynamical stability and have failed to match the accuracy of ANNs. Here we show how an SNN can overcome many of the shortcomings that have been identified in the literature, including offering a principled solution to the dynamical "vanishing spike problem", to outperform all existing shallow SNNs and equal the performance of an ANN. It accomplishes this while using unsupervised learning with unlabeled data and only 1/50th of the training epochs (labeled data is used only for a simple linear readout layer). This result makes SNNs a viable new method for fast, accurate, efficient, explainable, and re-deployable machine learning with unlabeled data.**

## I. Introduction

In recent years Deep Artificial Neural Networks (DNNs) have come to exceed state-of-the-art performance in many traditional Machine Learning (ML) domains, and have progressed in many domains that were previously inaccessible to ML approaches, such as face, image, video and speech recognition, and natural language understanding and generation. DNNs trained with gradient descent (GD) on an objective function in a process known as Deep Learning (DL) often excel at discovering complex features in their input data [1-3]. However DNNs also typically require exceptionally large numbers of free parameters and commensurately sized training datasets [3, 4], meaning that they are poorly suited for many real-world applications where the problems are not necessarily static or *a priori* easy to capture explicitly, where training data may be sparse or expensive to collect, high power data servers are not available, there are real-time or latency constraints, communication channels may be unreliable or open to attack, adaptation or online training may be required, privacy or security is a factor, or any number of other reasons why relying on large, remote, power-hungry and unadaptive ML models may be inappropriate [5-7].

An emerging alternative to DNNs is Spiking Neural Networks (SNNs). Rather than continuous activation functions, SNNs use discrete events (spikes) to transmit information between neurons, and rather than learning through global gradient descent, SNNs can learn using *local unsupervised correlation- or causation-based learning rules* such as spike timing dependent plasticity (STDP [8, 9]) which strengthens connections between causally connected neurons (i.e. when one neuron elicits a spike in another neuron). Amid growing recognition of the practical and computational limits of DL and of supervised learning in general, many eminent DL researchers suggest that the path to further substantial progress in AI is through unsupervised learning [2, 10]. While some DNN architectures, such as autoencoders, adversarial networks and their variants, implement a form of "self-supervised" learning, they still require definition of an objective function and they still rely on global error feedback for training, and so they continue to suffer many of the same disadvantages as fully supervised DNNs.

Both brains and SNNs have the capacity to compactly and efficiently represent information in the timing of their discrete spikes [11-15]. Nevertheless, many SNN architectures rely on spike-rate coding as a proxy for the continuous activation functions of DNNs, but in doing so they forgo the potential advantages of spike-time coding. Spike-time coding using sparse spiking events is exceptionally efficient in terms of both

This research is supported by the Commonwealth of Australia as represented by the Defence Science and Technology Group of the Department of Defence. *(Corresponding author: Peter G. Stratton).*

Peter G. Stratton was with University of Technology Sydney, Sydney, NSW 2007 Australia. He is now Associate Professor with the School of Electrical Engineering and Robotics, Queensland University of Technology, Brisbane Qld 4001, Australia (email: peter.stratton@qut.edu.au).

Andrew Wabnitz is with Defence Science and Technology Group, Department of Defence, Edinburgh, SA 5111 Australia (email: andrew.wabnitz1@defence.gov.au).

Chip Essam is with Cuvos Pty. Ltd. Sydney, NSW 2000 Australia.

Allen Cheung is an independent researcher.

Tara J. Hamilton was with University of Technology Sydney, Sydney, NSW 2007 Australia. She is now with Cuvos Pty. Ltd. Sydney, NSW 2000 Australia and Adjunct Associate Professor with UTS (email: tara.hamilton@cuvos.com.au).

Appendix proofs and Extended Data are available online.

energy requirements and information density, while simultaneously having high memory capacity and strong generalization ability [16-21]. Additionally, unsupervised local learning in SNNs is simple and fast, allowing them to learn more rapidly and with much improved data efficiency over DNNs [22, 23] (and also over SNNs that use approximations of gradient descent). Finally, dispensing with data labels, aside from a final simple linear readout stage if needed, means that collecting and curating training data is often quicker and cheaper than for DL, for which it can be an onerous task. We call SNNs that use spike timing to represent information, and that use unsupervised learning rules to train the weights: Spike-Timing Unsupervised Neural Networks (STUNNs).

Despite their potential advantages, STUNNs and SNNs in general remain on the fringe of ML research. Several factors have likely contributed to this failure to attract significant attention:

1. In multi-layer SNNs, spike propagation tends to fail as spikes pass between layers [24-27]; this has been called "vanishing forward-spike propagation".
2. There is a commonly perceived inadequacy of unsupervised learning for reliably solving logical NOT-like and XOR-like problems [27-29].
3. The lack of an objective function means performance is often below comparable DNNs.
4. Combining spike-time coding with unsupervised learning has proven challenging [1, 21, 22].

We show that the first problem (vanishing spikes) has potentially had widespread, severe and yet relatively unrecognized impact on the capacity of SNNs to function effectively. Our solutions significantly advance the state-of-the-art to the extent that a simple shallow STUNN using unsupervised local learning equals the performance of a shallow ANN on the MNIST task with only a fraction of the training. The solutions form a list of principles for robust unsupervised feature extraction using SNNs, and a new method for rapid and efficient machine learning with unlabeled datasets.

## II. METHODS

We implemented a spike timing unsupervised neural network (STUNN) that we called BLiTNet (Binned Linear Time Network). BLiTNet is an adaptation and extension of the Self Organizing Recurrent Network (SORN [30, 31]) to incorporate feedforward networks with multiple layers and to also incorporate sub-timestep resolution for the spike times. Like SORN, BLiTNet implements simple time-stepped neuron and synapse models that carry no state between timesteps (i.e. there are no membrane, synaptic or other time constants in the models). A timestep or time bin in the model is nominally equivalent to a gamma oscillation cycle in the brain. Gamma oscillations occur at 40-100 Hz and have been linked to active neural processing states, and the wavelength of 10-25 ms also approximates the time courses of neuron membranes and fast synaptic currents as well as spike timing dependent plasticity (STDP). This gamma-timestep abstraction allows the capture of active neural processing dynamics in a lightweight model suitable for digital hardware implementation.

### A. Core BLiTNet

In BLiTNet, neurons are organized into $N$ layers, typically with an input layer, an output layer and zero or more intervening (feature) layers. Connectivity between the layers may be full, sparse or spatially localized. While neurons in the brain are either excitatory or inhibitory, in BLiTNet we allow individual neurons to project both types of connections to other neurons, which is a simplification we employ to reduce the total neuron count. Connections cannot change type (i.e. a connection is either excitatory or inhibitory and cannot change sign).

An excitatory connection from neuron $i$ in layer $m$ to neuron $j$ in layer $n$ is given by $W_{ji}^{+nm}$ and an inhibitory connection by $W_{ji}^{-nm}$. The network state $x$ at time $t$ is given by the vectors $x^n(t)$ of spike amplitudes in [0,1] for all layers $n$ in $1..N$, where 0 represents no spike, 1 represents a full-strength spike, and values between 0 and 1 represent intermediate-strength spikes. In the brain, neurons that are more strongly excited fire earlier in each gamma oscillation cycle, and thus the amplitude of the input to each neuron is re-coded into the oscillation phase, allowing for a readout of amplitude within each individual gamma cycle [32]. So even though each timestep in BLiTNet involves only one network state calculation, we conceptualize spike amplitude as the timing of the spike within the timestep (i.e. spikes occur with sub-timestep resolution). An amplitude of 1 denotes the start of the timestep and 0 the end (Ext Fig S10). The network state $x$ therefore evolves as follows:

$$x_j^n(t+1) = \sum_{m=1}^N \sum_i x_i^m(t)\left(W_{ji}^{+nm} - W_{ji}^{-nm}\right) + \xi_j^n(t) + c_j^n - \theta_j^n \quad (1)$$

where $\theta$ is the neuron threshold, $\xi$ is a time-varying noise input uniformly distributed in $[0,\xi_{max}^n]$, and $c$ is a constant input to the neuron. The activity vectors $x$ are clipped in the range [0,1] where any $x > 0$ represents a spike. Initial thresholds are drawn from a uniform distribution in the range $[0,\theta_{max}^n]$, with $\theta_{max}^n$ usually set to 0.1 or 0.2.

Connections and thresholds are subject to plasticity rules. An excitatory connection $W_{ji}^{+nm}$ undergoes excitatory spike timing dependent plasticity (STDP) that strengthens the connection by a small amount $\eta_{STDP}$ when neuron $i$ fires in the timestep immediately prior to neuron $j$ firing, and weakens the connection by the same amount when neuron $i$ fires in the timestep immediately following neuron $j$:

$$\triangle W_{ji}^{+nm}(t) = \eta_{STDP}(t)\left[\Theta(x_i^m(t-1))\Theta(x_j^n(t)) - \Theta(x_i^m(t))\Theta(x_j^n(t-1))\right] \quad (2)$$

where $\Theta$ is the Heaviside step function. If a connection goes negative it is reset to a small positive value $\epsilon = 10^{-6}$. The STDP learning rate $\eta_{STDP}$ is initialized to $\eta_{init} = 0.001$ and then annealed during the $T$ training iterations according to:

$$\eta_{STDP}(t) = \eta_{init}\left(1 - \frac{t}{T}\right)^2 \quad (3)$$

The total excitatory weight to each postsynaptic neuron $j$ is normalized to a constant $k$ each timestep:

$$W_{ji}^{+nm}(t) \leftarrow \frac{k^{nm}W_{ji}^{+nm}(t)}{\Sigma_i \, W_{ji}^{+nm}(t)} \quad (4)$$

This normalization rule preserves the relative strengths of the connections while inducing competition between the connections since one connection can increase in efficacy only by decreasing the others. For small or simple networks, setting $k = 1$ is adequate, but for larger networks:

$$k^{nm} = (0.1/p^{nm})/\overline{f^m} \quad (5)$$

where $p^{nm}$ is the connection probability from layer $m$ to layer $n$ and $\overline{f^m}$ is the mean firing rate of neurons in layer $m$. Firing rate $f \in [0,1]$ is measured as the proportion of steps in which spikes occur. The normalization constant $k$ creates larger weights for both sparser connections and lower presynaptic firing rates, ensuring that postsynaptic neurons continue to use the full range [0,1] for spike amplitudes. The constant 0.1 sets the mean spike amplitude in layer $n$, and is used rather than 1 because the latter would cause half the spikes to be clipped at maximum amplitude.

Inhibitory connections undergo inhibitory STDP that strengthens the connection by $\eta_{STDP}$ when neuron $i$ fires in the timestep immediately prior to neuron $j$ (this rule is identical to excitatory STDP) and weakens the connection when neuron $i$ fires and then neuron $j$ does not fire in the timestep immediately following (this rule is different to excitatory STDP). Additionally, the weakening term is normalized by the firing rate of the postsynaptic neuron ($f_j$) to balance the strengthening and weakening terms for sparsely firing postsynaptic neurons:

$$\triangle W_{ji}^{-nm}(t) = \eta_{STDP}(t)\left[\Theta(x_i^m(t-1))\Theta(x_j^n(t)) - \Theta(x_i^m(t-1))\Theta(1-x_j^n(t))f_j^n\right] \quad (6)$$

If a connection goes positive it is reset to a small negative value $\epsilon = -10^{-6}$. Inhibitory connections are not normalized to a fixed value. Their role is to balance the excitatory inputs which can vary considerably depending on input statistics, connection densities, target firing rates, etc. Instead, all inhibitory connections to a postsynaptic neuron increase slightly when net input to that neuron is positive, and decrease slightly when net input is negative:

$$W_{ji}^{-nm}(t) \leftarrow W_{ji}^{-nm}(t)\left(1 - \eta_{STDP}(t).\Theta(\Sigma_i \, x_i^m(t))\right) \quad (7)$$

Intrinsic threshold plasticity (ITP) adjusts the firing threshold of each neuron such that the neuron reaches a desired target firing rate:

$$\Delta\theta_j^n(t) = \eta_{ITP}(t)\left(\Theta\left(x_j^n(t)\right) - f_j^n\right) \quad (8)$$

where $f_j^n$ is the desired firing rate of neuron $j$ in layer $n$. Unless otherwise stated, firing rates are set in a uniform distribution in [0.03,0.25]. If a threshold falls below 0 it is reset to 0 (i.e. neurons are not allowed to be spontaneously active). The ITP learning rate $\eta_{ITP}$ is initialized to $2\eta_{init}$ and then annealed during the $T$ training iterations according to:

$$\eta_{ITP}(t) = 2\eta_{init}\left(1 - \frac{t}{T}\right)^2 \quad (9)$$

*B. Spike forcing*

In some cases we used "spike forcing" of neurons in the output layer to create a supervised spiking readout of the result that is represented in the underlying feature layer, by forcing output neurons to spike or not spike as required. Spike forcing is similar to the delta learning rule and to standard STDP except there are more learning cases to consider. Since there are two types of spikes – network spikes (those that are driven by the network), and forced spikes (those that are pre-specified) – there are four spiking conditions. A neuron could have:

1. a forced spike and no network spike
2. a network spike and no forced spike
3. both forced and network spikes
4. neither forced nor network spikes

For training the neuron thresholds $\theta$, the learning rules ensure that each neuron sets its threshold to spike when a forced spike is desired and not spike when a forced spike is not desired (Table I):

TABLE I
SPIKE FORCING THRESHOLD LEARNING RULES

| Condition | 1. Forced only | 2. Network only | 3. Both | 4. Neither |
|---|---|---|---|---|
| $\Delta\theta$ | $-\eta_{STDP}(t)$ | $\eta_{STDP}(t)$ | 0 | 0 |

For training the connection weights $W$, the learning rules strengthen connections from input neurons that should cause output spikes, and weaken connections from input neurons that should not (Table II):

TABLE II
SPIKE FORCING WEIGHT LEARNING RULES

| Condition | 1. Forced only | 2. Network only | 3. Both | 4. Neither |
|---|---|---|---|---|
| $\triangle W_{ji}^{+nm}(t)$: (excitatory connections) | **Eq. (2)** i.e. normal STDP on forced spikes | **-(Eq. (2))** i.e. anti STDP on network spikes | **Eq. (2)** i.e. as normal | 0 i.e. as normal |
| $\triangle W_{ji}^{-nm}(t)$: (inhibitory connections) | **-(Eq. (6))** i.e. anti STDP on forced spikes | **Eq. (6)** i.e. normal STDP on network spikes | **Eq. (6)** i.e. as normal | 0 i.e. as normal |

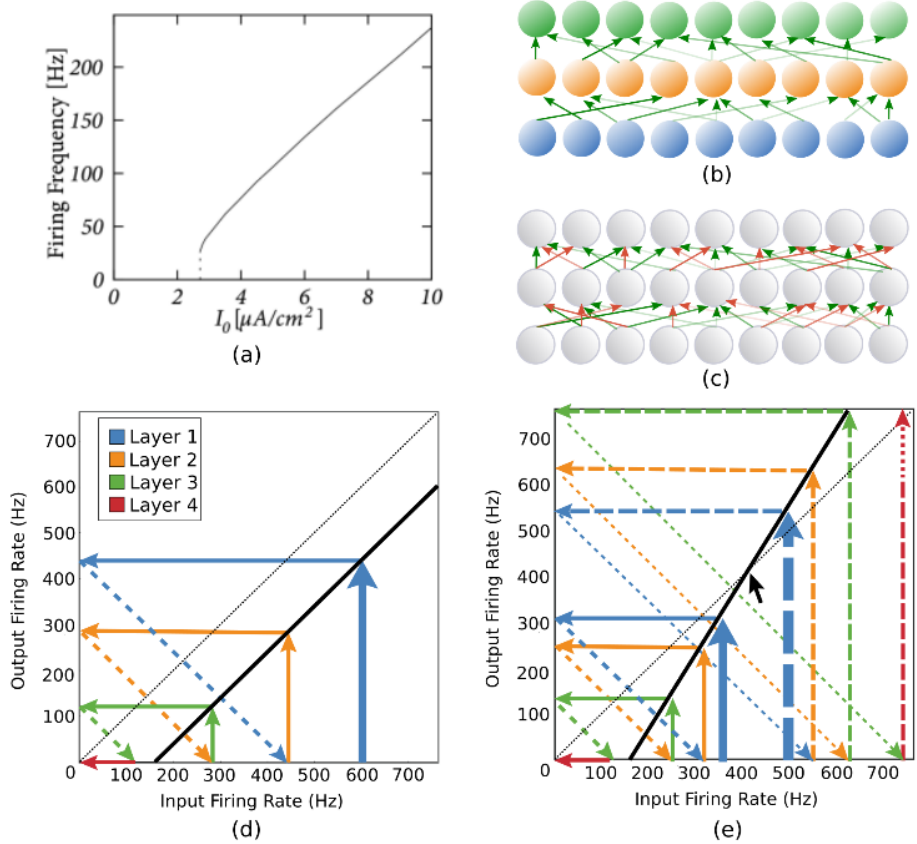

**Fig. 1.** Firing thresholds of spiking neurons and networks. (**a**) Firing frequency against input current for the Hodgkin-Huxley neuron model, using input currents within physiological range [https://neuronaldynamics.epfl.ch/online/Ch2.S2.html]. (**b**) Network of forward-connected layers of neurons. (**c**) An alternative feedforward architecture with additional inhibitory connections that approximately balance the excitatory connections. (**d**) Network gain function where gain = 1 invariably leads to vanishing spikes because of the positive threshold ($x$ intercept). (**e**) For network gain > 1, activity will either vanish (solid arrows) or saturate (dashed arrows) from the stable firing rate (small black arrow) depending on finite-size fluctuations.

*C. Signal propagation networks*

Parameters were as specified above and in the main manuscript, and as noted below.

- For all networks: $N = 10$, $T = 10000$, $\xi^1_{max} = 0.1$ and $\xi^{2..N}_{max} = 0$ (i.e. noise drives the input layer only).
- For networks with no plastic inhibition: $\eta_{STDP} = 0$.
- For networks with plastic inhibition: $\eta^+_{STDP} = 0$ and $\eta^-_{STDP}(0) = \eta_{init}$ (i.e. no plasticity for excitatory connections).
- For all networks that have equal target firing rates for all neurons: $f = 0.1$.
- For the last network with a range of target firing rates: $f$ uniformly distributed in [0.025,0.175].
- For networks with constant input: $c$ uniformly distributed in [0,0.099].

*D. Logical function networks*

The network architectures were as shown in the main manuscript. For all networks, $T = 10000$, $\xi = 0$, $\eta^+_{STDP}(0) = \eta_{init}$ and $\eta^-_{STDP}(0) = 0$ (i.e. no plasticity for inhibitory connections).

*E. MNIST networks*

Parameters were as specified above and in the main manuscript, and as noted below. For all networks: $N = 3$, $T = 60000$, $\xi = 0$ and $c = 0$.

All networks used 784 input neurons in the first layer, then variable numbers of feature-layer and output-layer neurons. The feature layer was trained first, then its learning rate was set to zero and the output layer was trained. Output neurons were split into 10 groups of equal size (one group for each digit) with each output neuron sampling a different subset of neurons in the feature layer due to sparse excitatory connectivity. Output neurons were trained with spike forcing. The weight matrices between layers were as follows:

$W^{+21}$: Full or spatially localized plastic connections as shown in manuscript.

$W^{-21}$: Sparse plastic connections with connection probability 0.015.

$W^{+32}$: Sparse plastic connections with connection probability between 0.2 and 0.3.

$W^{-32}$: Full or nearly full plastic connections with connection probability between 0.7 and 1.0.

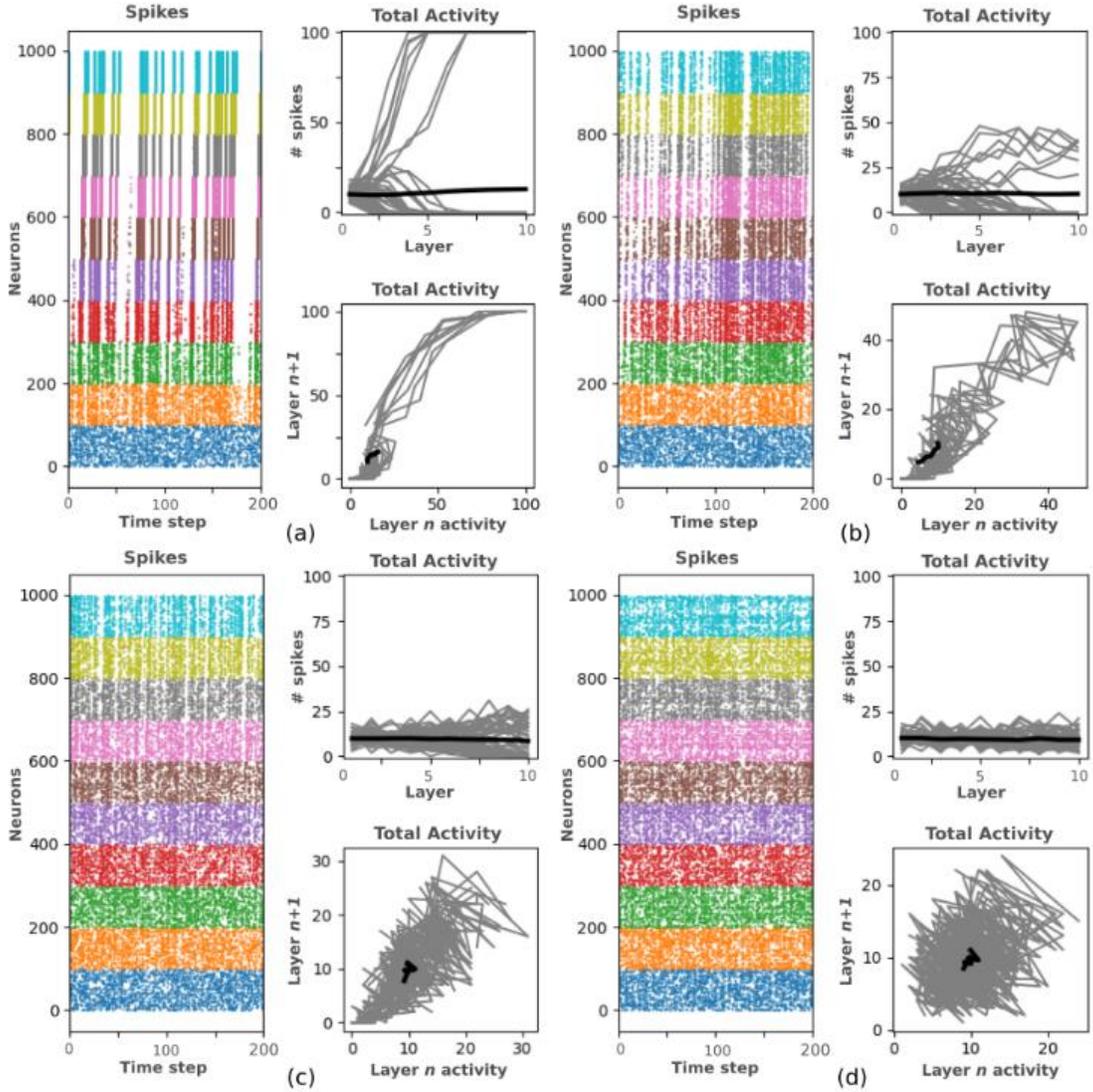

**Fig. 2.** Signal propagation through SNNs. (**a**) Failure of signal propagation through a standard feedforward SNN. Left: Spike raster plot over the last 200 testing steps. Neurons in the first (bottom-most) layer received random noise input. Activity always either vanished or avalanched. Top right: Number of neurons firing as activity propagated through each layer for the first 50 timesteps shown in Left (black = mean). In all cases activity either fell to zero or increased until every neuron was firing. Bottom right: Plotting number of neurons firing in layer $n$ at time $t$ against number firing in layer $n+1$ at time $t+1$ shows a tight correlation, r=0.900 (black = mean). (**b**) Fixed balancing inhibitory connections somewhat rescued the signal propagation. Left: Spike rasters showed much improved propagation but with many interspersed cases of propagation failure remaining. Top right: Spike avalanches were controlled, never involving all neurons at once. The mean firing rate was maintained by ITP despite that in cases when a signal vanished no further propagation was possible. Bottom right: However, the correlation of firing rates between layers remained very high, in this case r=0.920. (**c**) STDP of the balancing inhibition and constant input to each neuron resulted in much improved signal propagation and a significant reduction in signal correlation, r=0.505. (**d**) All mechanisms together (inhibitory STDP, constant input to each neuron, and different target firing rates for each neuron) resulted in almost perfect balance and propagation, reducing signal correlation to r=0.111.

After training on the 60000 MNIST training digits, network performance was assessed in two ways:
1. The output neuron group with the most spikes was used to classify each of the 10000 test input digits.
2. A decoder was trained using linear regression on the feature layer for each training digit and was then used to classify each test digit.

*F. Code availability*

Code will be made available for non-commercial use.

## III. RESULTS

Our results are presented in three sections below. The first shows the etiology and resolution of the vanishing spike problem, or "spike propagation failure". The second shows how the same solutions facilitate unsupervised learning of logic functions such as NOT and XOR. The last shows how the combination of these solutions improves state-of-the-art performance for shallow unsupervised learning of MNIST images.

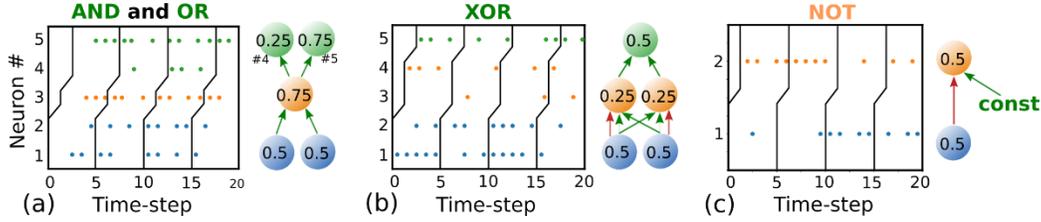

**Fig. 3.** A STUNN can solve logic functions (input neurons: blue; features: orange; outputs: green). Activity in each layer is delayed by one timestep from the previous (vertical black lines). Target firing rates are shown inside each neuron. (**a**) AND (neuron #4) and OR (#5) can be learned from a single feature neuron since this neuron fires earlier in its timestep (i.e. more strongly; see Methods) when driven by both input neurons simultaneously. (**b**) By adding sparse inhibitory balance to the network (red arrows) and setting lower firing rates for the feature neurons, XOR is learned by the smallest network possible. (**c**) By further adding constant input to a feature neuron, NOT is learned.

*A. Solving Spike Propagation Failure*

Neurons are *leaky* integrators of their input currents, which makes their response to constant input highly nonlinear and even discontinuous. An input current to a neuron that is smaller than the neuron's maximum leak current, even if applied indefinitely, will never lead to an output spike. With an input current just slightly larger than this specific *threshold* current, a typical neuron will spike rhythmically, for as long as the input is applied, with a period determined in large part by the time constant of the threshold leak current. Therefore there is a discontinuous jump from no spiking to spiking continuously at a constant low rate. For input currents substantially larger than the threshold current, a neuron enters an approximately linear regime that is affected less and less by the leak as the magnitude of the input current increases. For even higher input currents, a neuron's refractory period (i.e. the time following a spike during which a neuron is unable to fire again) begins to dominate and the firing rate begins to saturate. Beyond this level the firing rate plateaus, however this level of input is typically non-physiological and if sustained will usually lead to neuron death due to excito-toxicity. A typical single-neuron gain function is shown in Fig 1a.

The single-neuron gain function can be extended to a network of interconnected neurons where neurons are organized in layers, and where neurons in each layer are unidirectionally connected to (a subset of) neurons in the next layer (Fig 1b). For the same reason that a single neuron requires input current beyond a threshold in order to produce any output spikes, a network layer requires a certain number of input spikes from the previous layer to produce any output spikes of its own. A schematic plot of the network gain function showing the output layer firing rate against the input layer firing rate is analogous to the previous single-neuron gain function (Fig 1d and Extended Fig S1). The mapping from input layer population firing rate to output population firing rate (Fig 1d, dark black trace) shows the effect of the non-zero spike firing threshold. If the population firing rate of the first (input) layer begins at 600 Hz (bold arrow), the subsequent rate of the second layer will be lower, as shown by following the arrows. When this activity is then used as the input to the next layer, the resulting activity level is even lower. Activity entirely vanishes after propagating through several layers.

If the network gain is increased to greater than 1 (Fig 1e and Extended Fig S2), then there will be one input firing rate for which the input and output firing rates will be exactly equal (small dark arrow). However, due to the quantized, all-or-nothing spike currents and the finite number of neurons in each layer (i.e. finite size effects), it is impossible to maintain this exact rate. Small variations from one moment to the next will accumulate. If the variation is downwards then on the next layer the downward drop will be magnified (due to gain > 1) and once again activity will quickly vanish (Fig 1e, solid arrows). Similarly, if the variation is upwards then this upward tendency is also magnified and activity will evolve into unconstrained spike avalanches where every neuron spikes continuously (Fig 1e, dashed arrows).

We simulated a network consisting of 10 layers of 100 neurons per layer, sequentially connected by random fixed-weight feedforward connections with 10% connection probability. Neurons in the bottom layer received random noise input. Every neuron used intrinsic threshold plasticity (ITP; see Methods) to reach a target firing rate of 0.1 (i.e. firing in 10% of the time-steps). As spiking activity propagated through the layers, input patterns that randomly contained fewer spikes than average were further under-represented in subsequent layers, while patterns with randomly more spikes provoked inordinately large responses, resulting in most of the network's spikes being elicited for only the largest input patterns (Fig 2a, left and top right panels). Even in just the propagation of activity from the first layer to the second, magnification of variations in activity levels are clear (i.e. input patterns that were randomly smaller (/larger) in the first layer recruited even less (/more) activity in the next layer). The number of neurons firing in each layer was strongly correlated with the number in the layer below (r=0.900, Fig 2a bottom right panel).

Note that this signal propagation failure is independent of the neuron model being used; it applies from the simplest linear threshold unit all the way to the most biophysically detailed models, and indeed to real neurons. The only requirement is for the neuron to have a non-zero threshold and rectified output (see our novel analysis: Appendix). The default behaviour of SNNs is therefore to tend towards dynamical quiescence or saturation which, after spike propagation through several layers, diminishes their information-carrying capacity to essentially zero. This effect has been termed "vanishing forward-spike propagation" [1]. We name it "spike propagation failure" since the failure can be either quiescence (vanishing) or saturation (avalanche) [24-27]. The problem has been blamed on "inappropriate weights and thresholds" [33], but it is clear from the above that no weight settings could alleviate the problem, since these will only change the overall gain, while removing the threshold nonlinearity will severely impact both the stability and computational capacity of the networks, since the system would be reduced to linear.

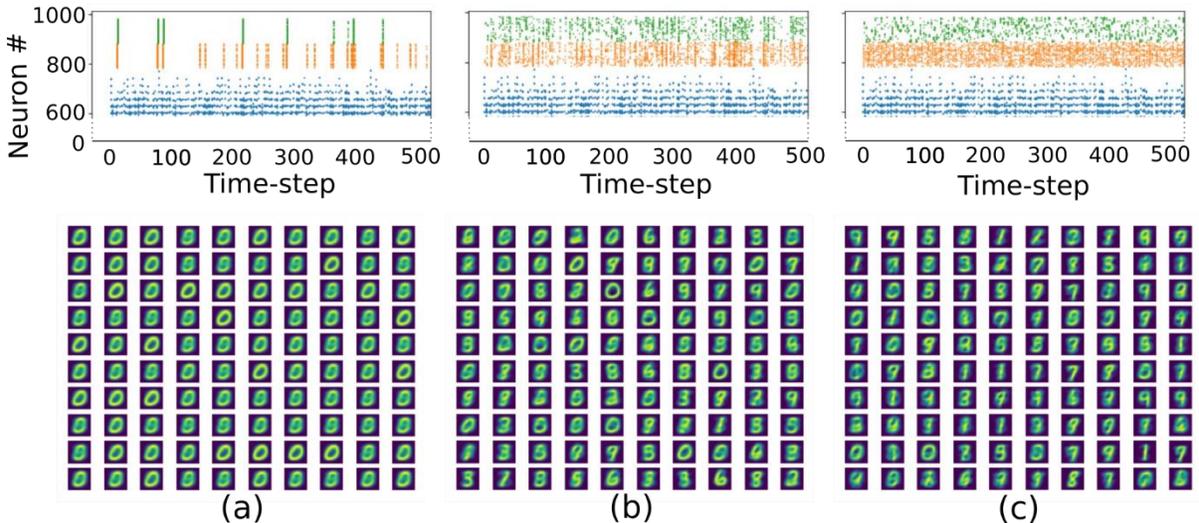

**Fig. 4.** Balancing inhibition allowed the network to reliably and robustly extract features from MNIST data. (**a**) No inhibition. (**b**) Fixed balancing inhibition. (**c**) Plastic balancing inhibition. **Top row:** Spike raster plots (input spikes: blue (not all neurons shown); feature spikes: orange; output spikes: green). Without balancing inhibition between layers (a), the feature and output spikes occurred only as rare bursts. With fixed balancing inhibition (b), feature and output spikes remained somewhat sporadic and nonuniform. Only plastic balancing inhibition (c) resulted in reliable signal propagation through the layers. **Bottom row:** Learned connection weights (features) from the input layer to the feature layer. Those digits that are comprised of the fewest pixels (i.e. the smallest inputs, like '1' and '7') were only well-represented by the network with plastic balancing inhibition (c).

A general solution to this problem within the ML literature has not been forthcoming. Apart from the few references cited above, there is surprisingly little recognition of the problem despite it being ubiquitous in all SNN models. The failure to recognize this issue, or to take it seriously enough, may have been at considerable detriment to SNN research.

However, a basic solution is immediately available and is an important component of theories behind cortical dynamics and criticality in the brain. The idea is to use excitation/inhibition (E/I) balance; that is, that excitatory and inhibitory connections, and therefore dynamic synaptic currents, can be closely matched (Fig 1c). Within the context of recurrent SNNs and cortical dynamics, E/I balance ensures that network activity remains in a chaotic or critical regime since spikes are driven by small fluctuations in balance that are highly dependent on initial conditions and ongoing inputs [26, 27, 30, 34]. In the context of spikes propagating through a feedforward network [24, 25], balance removes the dependence of the activity level of layer n+1 ($a_{n+1}$) on the activity level of layer n ($a_n$). Therefore, rather than depending on the absolute magnitude of $a_n$, $a_{n+1}$ depends only on small deviations from the balance. The imperfections in balance can be spatial (connections that are sparse, random, or in some other manner not fully overlapping) and temporal (a delay between the arrival of excitatory and inhibitory signals). Due to the approximate E/I balance, the mean input to each neuron over time is close to zero rather than being positive, restoring the linearity of the network gain without upsetting the stability. In this regime, individual neurons can still have positive thresholds which allow them to fire sparsely rather than firing every time their inputs fluctuate above zero.

Fig 2b shows results from a network that used balancing inhibitory connections as shown in Fig 1c. Signal propagation was dramatically better but still imperfect. To further improve the propagation, three additional mechanisms with inspiration from neuroscience [35, 36] were considered:

1. Inhibitory STDP [35] (i.e. plasticity of the balancing inhibition).
2. Constant input to each neuron [36].
3. Different target firing rates for each neuron.

When applied individually, each of the three mechanisms made little difference to the signal correlation, at best reducing it to about r=0.85. However, when testing pairs of mechanisms, the pairing of inhibitory STDP and constant input made considerably more difference, reducing correlation to 0.505 (Fig 2c). The remaining pairings (different firing rates with respectively STDP and constant input) made little further improvement beyond the improvement using each mechanism individually.

However, when all three mechanisms were applied together, the balance was much further improved, signals propagated without obstruction (Fig 2d), and signal correlation reduced to 0.111. Being driven predominantly by small fluctuations in the E/I balance, the network gain function was discontinuous (Fig 2d, bottom right).

*B. Solving XOR and NOT*

The exclusive-OR function (XOR) has a long history in the neural network field. During the 1940's, 50's and 60's the Perceptron (network of linear threshold units) was investigated as a model of neural function, and interest in neural computation was flourishing. Then in the late 60's it was shown that the simple XOR problem, being linearly inseparable, could never be solved by a single-layer perceptron, and no training algorithms existed for multi-layer perceptrons (MLPs). This realization contributed to the first "AI winter", which for neural networks lasted until just after 1986 when it was shown that with a few clever techniques the chain-rule backpropagation algorithm could be applied to train MLPs, and the XOR problem could be solved by neural networks using *supervised* learning.

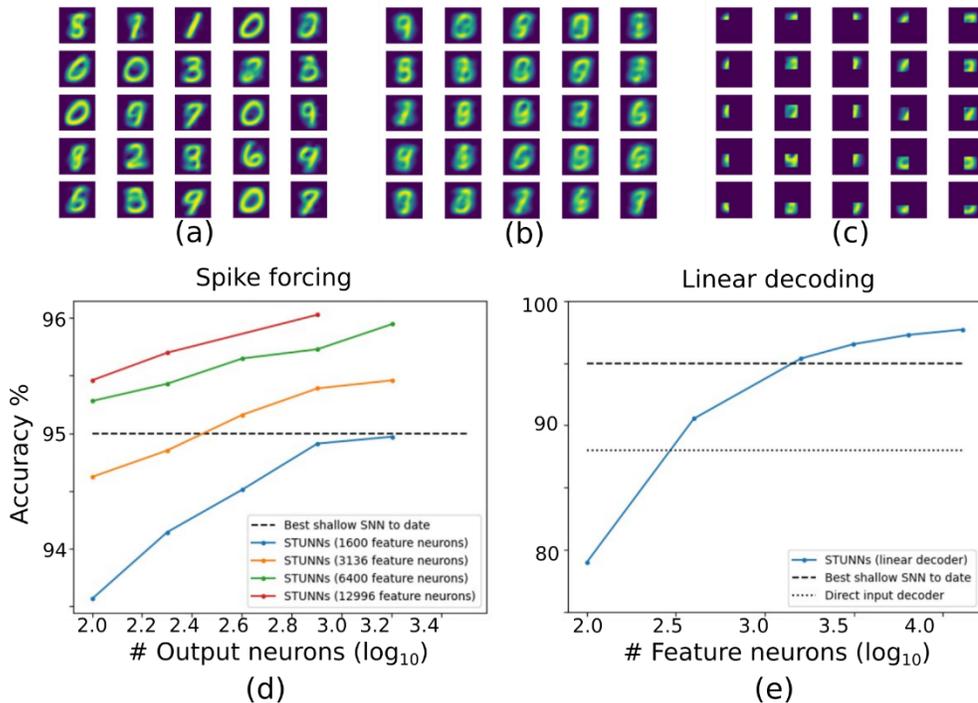

**Fig 5:** Effect of firing rates and numbers of neurons on features and performance. **a.** Low firing rates encouraged the formation of very specific features (e.g. single digits). If there are a limited number of neurons then some features are omitted. **b.** Conversely, high firing rates encouraged combinations and superpositions of similar features, giving more complete coverage of the feature space and a more distributed neural code. **c.** The combination of a distribution of firing rates from low to high, with spatially localized connections, resulted in a range of small features being discovered and the best network performance overall. **d.** Performance of the STUNN using spike forcing of the output neurons (max. 96.0%). **e.** Performance of the STUNN using linear decoding of the feature neurons (max. 97.7%).

Notwithstanding the kernel trick (where random projections from the input to a higher-dimensional feature space result in at least one dimension upon which features can be linearly separated [37]), an implicit assumption in the field that seems to dominate to this day is that *unsupervised* learning will struggle to reliably learn XOR and similar linearly inseparable problems, since there is no impetus (error signal or gradient) for the network to follow in order to construct the required representations (but see [28, 29]). Indeed, even to solve something as simple as the NOT function, implemented in an SNN with one input neuron and one output neuron, would require the output neuron to remain silent when the input neuron emits a spike, and for the output to spike when the input is silent. Clearly for any purely input-driven network this is not possible, and it is unclear how any unsupervised learning algorithm could learn this function [27].

Here we show that *minimally sized* SNNs, using combinations of network balance, constant input, and a distribution of feature neuron firing rates (the same mechanisms that permit stable signal propagation), can trivially learn unsupervised spike temporal coding versions of all logic functions (Fig 3). For example, NOT can be solved by a feature neuron that receives an *inhibitory connection* from an input neuron, causing it to *not emit* a spike when the input neuron spikes, in combination with a *constant input* that causes it to *emit* a spike when the input neuron does not (Fig 3c). The learning in this case entails the neuron finding a suitable firing threshold that lies between the constant input and the constant input less the inhibitory connection weight. For this learning to work effectively the neuron needs to have a target firing rate that is the complement of the input neuron's rate, and then using intrinsic threshold plasticity (ITP) the neuron will converge onto a suitable threshold. In this basic example we simply choose a suitable target firing rate that matches the logical function result that we desire. In more complex cases, larger networks with a distribution of target firing rates over the population of feature neurons, in combination with sparse random connections (so that some neurons receive only excitatory connections from given inputs, some receive only inhibitory, some both, and some neither), can learn all logic functions simultaneously. Temporal coding allows the networks to solve these functions using the minimum possible number of spikes.

*C. Solving MNIST with performance equalling ANNs*

Using our Binned Linear-Time Network (BLiTNet) implementation of a STUNN, we trained a network using the MNIST dataset (784 input neurons) fully connected with plastic excitatory connections to 100 feature neurons (see Methods for details). With no inhibition in the network the results on the MNIST test set were very poor (Fig 4a). The initially imperfect network balance was worsened by the plasticity in a vicious feedback loop where the network would respond with more spikes to those digits that had more active pixels, i.e. more input spikes, and the plasticity would enhance that response even further, such that almost every feature neuron ultimately learned to respond only to the largest digits (0 and 8). Most SNN models incorporate some form of inhibition, and adding fixed balancing inhibition gave significantly improved results as expected, but the balance was still imperfect and many input digits failed to elicit any output spikes due to signal propagation failure (Fig 4b). However, with the addition of plasticity to the inhibition the network achieved excellent balance, and each feature neuron learned to respond to a different digit or in some cases superimposed combinations of several similar digits (Fig 4c). Those digits that were made up of the fewest pixels (i.e. the smallest inputs, such as '1') were well-represented only by this network. The plastic inhibition learned to approximate the negation of the excitatory connections (Extended Fig S4). The specific characteristics of the excitatory features that were learned were strongly influenced by the target firing rate of each feature neuron (Fig 5a, b and Extended Fig S5). For this reason, a distribution of firing rates across the neurons in the feature layer allowed the network to reliably capture a broad range of features that occurred with different frequencies in the input.

We tested many networks with different numbers of neurons in the feature and output layers, and different connection sparsities (Extended Fig S6 and S7). We found that using spatially localized receptive fields with plastic balancing inhibition and a range of firing rates for the feature neurons gave optimum performance (Fig 5c and Extended Fig S8). Results are summarized in Fig 5 (d and e). Previous best performance on MNIST using an unsupervised shallow SNN was 95.0% accuracy on the test set [38]. Using spike forcing of the output layer (i.e. a spiking readout of the features in the feature layer – see Methods) we achieved 96.0% accuracy using either 12996 feature neurons with 800 output neurons, or 6400 feature neurons with 1600 outputs (Fig 5d). Additionally, using a simple linear decoder on the feature layer with 12996 neurons we achieved an astounding 97.69% accuracy. For comparison, a shallow ANN trained with GD using a three-layer architecture (784–128–10 units) achieves 97.6% accuracy after 5 epochs and 98.0% after 50 epochs. Adding units to the feature or output layers of the ANN does not improve performance. Recently an SNN trained with GD reached 97.2% after 40-50 epochs [39]. Our STUNN compares favourably to both, with 97.7% accuracy after only 1 epoch (continued training results in little or no further improvement).

IV. DISCUSSION

The problem of signal propagation failure in feedforward SNNs is ubiquitous across all spiking neuron models as well as in real neural networks in the brain. The issue is rarely recognized in ML research using SNNs, and to the best of our knowledge its root cause has not been formally identified. We have shown analytically that the problem is caused by the nonlinear rectified neural gain function (see Appendix), and shown numerically that it severely impacts the ability of spikes to propagate reliably through even a very small number of neural layers. The solutions we have presented provide a list of principles for reliable signal propagation and for facilitating robust feature extraction in SNNs:

1. Balancing inhibition compensates for the nonlinearities in the neural gain function.
2. Inhibitory plasticity compensates for non-uniform input statistics.
3. Constant input prevents random-walk dynamics from hitting zero (i.e. from "falling off the edge")
4. Firing rate distribution prevents spike correlations caused by over-synchronization.

All these principles have their roots in neuroscience. While balancing inhibition has been used in ML applications before [29, 30, 37], we have shown that all four mechanisms are required to maintain near-perfect signal propagation. The cortex of the brain is known to operate in a state of dynamic balance between excitation and inhibition, and learning is known to engage inhibitory plasticity to maintain the balance. Cortical neurons also receive constant input from the brainstem (via the thalamus), the magnitude of which controls the dynamic cortical state and without which the cortex shuts down. Finally, cortical neurons display a broad range of firing rates. It is reasonable to propose that the brain uses these mechanisms for purposes similar to those shown here: for maintaining dynamical stability so that signals propagate reliably, and for improving functional computational outcomes such as feature extraction and logic calculations. While it may seem self-evident that improving the network dynamics will also improve computational function, the effectiveness of firing rate distributions for matching input feature frequencies, and of constant inputs for learning nonlinear logic functions, was unexpected. Note that perfect balance implies that firing rates from one layer to the next are uncorrelated, which would render spike rate coding inoperative, contrary to the temporal coding used here. Note also that perfect balance may not always be desirable; slight imperfections can potentially be used to route information through different neural circuits (Extended Fig S3).

The demonstration of unsupervised spike-time learning of logic problems including NOT and XOR, using minimal networks and minimal numbers of spikes, means that unsupervised learning in general, and STUNNs in particular, are applicable to a whole new class of problems. We showed how STUNNs can learn to solve logic functions of two inputs. For more than two inputs, there is clearly an explosion in the number of possible input combinations ($2^n$), but this is offset by the exponential explosion in the number of possible combinations of learned feature representations and their logical complements (i.e. NOT(feature)). These combinatorial codes are possible with the sparse temporal codes used in this work but are not supported by the dense rate codes used elsewhere [28], unless networks are pre-wired into clusters [29] or otherwise hand-tuned [27]. However, the generalization of XOR to bitstrings of arbitrary length is called the bit parity problem, and while there is a simple procedure to calculate it sequentially, how to *learn* a general solution is still a topic of active research [14]; even supervised learning cannot currently solve bit parity for arbitrary length strings.

The brain contains many neurons that hold conjunctive representations [40], and even bees have the ability to represent zero or the absence of a stimulus [41], although it has been unclear how. STUNNs with constant input to each feature neuron develop a specific, unique and identifiable feature-layer representation of an empty input layer, which helps explain how this feat could be accomplished by neural circuits (Extended Fig S9). Note that constant input was not used in the MNIST network since many MNIST pixels around the edges of the digits are always off, meaning that those feature neurons that developed representations to these blank regions were not helpful for identifying digits and were essentially wasted (Extended Fig S9). While the network outputs easily learned to ignore those specific neurons, overall performance suffered slightly due to there being effectively fewer useful feature neurons that actually represented the digits.

An important concept in STUNNs ties together the closely related ideas of 1. sparse excitatory and inhibitory connections contributing to transient departures from perfect balance; 2. a distribution of properties like firing rates and constant inputs to each neuron and; 3. the subsampling of the input space that is a consequence of the previous two ideas. Balance must be perfect or nearly so *on average*, but it is also important that it is imperfect for any given input in order for some neurons to fire while others are suppressed. Similarly, there should be a good chance of a neuron with a given target firing rate connecting to a certain subset of inputs for detecting a feature of matching frequency on those inputs. Likewise, neurons with a constant input need to receive stronger inhibitory than excitatory input from a feature or features in order to calculate NOT(feature/s). Sparse connections, a distribution of neuron properties, and a large number of feature neurons, are instrumental in ensuring that all this can occur. This is what we see in the brain and, fortuitously, this distribution of properties is perfect for implementation in cheap neuromorphic hardware because elements do not need to be constructed to be exactly alike with high precision. Instead, natural variations in hardware characteristics can be exploited in service of the learning algorithms [42, 43].

STUNNs, and unsupervised learning in general, need to discover all the available correlations/features in the data, because with no objective function gradient to follow there is no indication of which features will be required to solve any given problem. In contrast, DNNs will only develop representations of features that are useful for the given problem (i.e. that follow the gradient). For this reason, STUNNs need to be larger than DNNs for solving comparable tasks, at least during training. However this need not be a disadvantage:

- First, since learning is local and representations are sparse, training speed is limited only by how long it takes to get a good sample of the input distribution, and not by global dependencies. Training is therefore lean and fast.
- Second, STUNNs can potentially be pruned after training to retain only those features that are most useful for the final readout layer (if one is needed) or that have strong connections to downstream processing stages.
- Third, labels are never required for training, only for the final readout, so training data is usually easy and economical to collect.
- Fourth, STUNNs can be re-tasked simply by retraining or recalculating the output layer, in contrast to DNNs. This capability should allow for near-flawless transfer learning, which we will investigate in future work.
- Fifth, STUNNs generate intuitive, explainable, parts-based features, somewhat akin to non-negative sparse codes [16], unlike DNNs for which features and decision boundaries are often counter-intuitive [44] rendering them difficult to interpret and prone to subtle and difficult-to-detect adversarial attacks. We leave detailed investigation of this characteristic of STUNNs for future work.

It should be clear that STUNNs are more suitable for learning scenarios where conventional GD for DNNs is limited by power, computational resource usage, training time, data collection and/or labeling requirements, lack of explainability or lack of transferability. While GD can lead to superhuman performance on certain well-constrained tasks for which a single objective can be explicitly defined, supervised methods are often brittle and highly dependent on the training dataset. For more general and difficult-to-define learning scenarios encompassing most real-world tasks, unsupervised learning will likely prove to be more generally useful. STUNNs use unsupervised local learning to swiftly build a complete, efficient dictionary of the input space, making them the best architecture to date for performing rapid, low-power machine learning using unlabeled data.

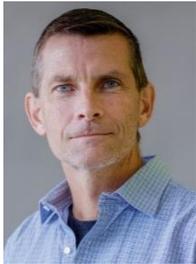

**Peter G. Stratton**
Queensland University of Technology, Australia.
https://orcid.org/0000-0002-3312-7505

Peter Stratton received his PhD in computer science and neural networks from The University of Queensland, Australia in 2002. He is currently an Associate Professor in the School of Electrical Engineering and Robotics at the Queensland University of Technology, Brisbane, Australia, and a Principal Research Fellow in neuromorphic engineering. He was previously a Research Fellow in neuromorphic algorithms at the University of Technology Sydney. His neuroscience background comes from nearly 10 years at the Queensland Brain Institute. His primary research interest is to understand the computational principles that are implemented by nervous systems, and to apply these to complex engineering problems in information processing and robotics.

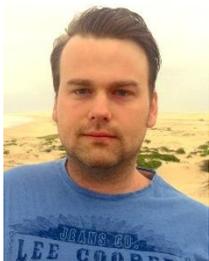

**Andrew Wabnitz**,
Defence Science and Technology Group.
https://orcid.org/0000-0002-2038-4174

Andrew Wabnitz received the B.E. (Hons) degree in electrical engineering from the University of Adelaide, Australia, in 2005, and the Ph.D. degree from the University of Sydney, Australia, in 2013. Andrew has over 10 years experience across industry, academia and defence working on projects involving embedded system design, software development and FPGA design for applications in biomedical devices, cyber security and space. He is currently a senior researcher of Cognitive Technologies in the Defence Science and Technology Group, Department of Defence, Australia. His primary research interests include neuromorphic computing, bio-inspired algorithm design, and the application of this technology to real-world problems.

**Chip Essam**, photograph and biography not available at the time of publication. https://orcid.org/0000-0003-4091-5304

**Allen Cheung**, photograph and biography not available at the time of publication. https://orcid.org/0000-0001-9770-217X

**Tara J. Hamilton** (Member, IEEE)
Cuvos Pty. Ltd.
https://orcid.org/0000-0003-2630-7011

Tara Julia Hamilton is currently the Principal Scientist at Cuvos and an Adjunct Associate Professor at the University of Technology Sydney (UTS). Her research interests include: analog and mixed-signal integrated circuit design, neuromorphic systems, and bio-inspired machine learning. Tara is the author of over 120 research papers and 3 international patents. Tara works with leading health, defence, and technology companies and she is focused on developing innovative solutions to real-world problems.

# Appendix

Fig A1 shows a schematic diagram for recording a neuron's membrane potential.

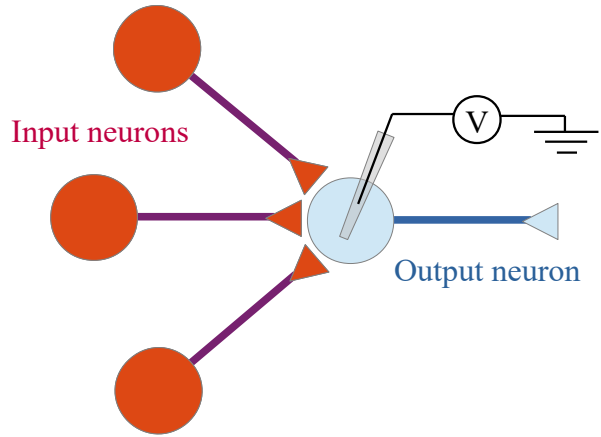

Fig A1: Schematic diagram illustrating hypothetical membrane potential recording from an output neuron in a feed-forward spiking neural network model.

## 1. RC circuit model of a leaky integrate-and-fire (LIF) neuron

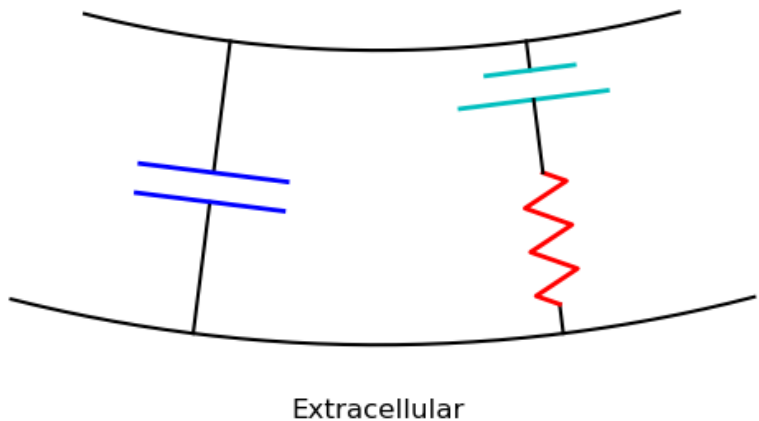

Fig A2: RC circuit model of the neuronal membrane.

Based on the equivalent RC circuit model of the neuronal membrane (Fig A2), the governing equation for membrane potential dynamics is

$$\tau_m \frac{dV(t)}{dt} = -(V(t) - V_{rest}) + RI(t) \qquad (1)$$

where $\tau_m = RC$ is the membrane time constant, $R$ is the electrical resistance of the membrane, $C$ is the charge capacitance of the membrane, $V(t)$ is the membrane potential at time $t$, $V_{rest}$ is the resting membrane potential, and $I(t)$ is the total current passing across the membrane at time $t$.

## 2. Average current is proportional to input frequency

For a regular input spike train of frequency $\lambda_{in} = 1/\Delta t$, we can compute the average input current.

Consider a general synaptic current arising from one input spike:

$$I_1(t) = \sum_j a_j t^k e^{-t/\tau_j} \tag{2}$$

where $a_j$ is a constant which weights the gamma distribution component with time constant $\tau_j$, and $k \geq 0$. The synaptic current model described by (2) generalizes commonly used examples from the SNN literature. For example, if $k=0$, $a_1 = I_0$ and $a_2 = -I_0$ then the current profile for a single input spike occurring at $t = 0$ is given by $I_1(t) = I_0(e^{-t/\tau_1} - e^{-t/\tau_2})$ (REF). Another example is $k=1$ and $a = we/\tau$ in which case $I_1(t) = (wt/\tau)e^{1-t/\tau}$ (NEST appendix, Jordan et al).

Suppose the most recent spike of a regular input train occurred at $t = 0$. To find the average total current between $t = 0$ and $t = \Delta t$, it is necessary to sum the contributions from all previous spikes. Assuming linear summation of successive currents, the steady state average current is

$$\begin{aligned}
\bar{I}_\infty &= \frac{1}{\Delta t} \int_0^{\Delta t} I_\infty(t) \, dt \\
&= \frac{1}{\Delta t} \int_0^{\Delta t} \left( \sum_{j=0}^\infty I_1(t + j\Delta t) \right) dt \\
&= \lambda_{in} \int_0^\infty I_1(t) \, dt \\
&= \lambda_{in} \int_0^\infty \left[ \sum_j a_j t^k e^{-t/\tau_j} \right] dt \\
&= \lambda_{in} \sum_j \left[ -a_j \tau_j^{k+1} \Gamma(k+1, t/\tau_j) \right]_0^\infty
\end{aligned} \tag{3}$$

where $\lambda_{in}$ is the input spike rate, and $\Gamma(\alpha, \beta) = \int_\beta^\infty t^{\alpha-1} e^{-t} \, dt$ is the (upper) incomplete gamma function. Since $\Gamma(\alpha, 0) = \Gamma(\alpha)$ and $\Gamma(\alpha, \infty) = 0$ we can write (3) as

$$\bar{I}_\infty = \lambda_{in} \sum_j \left[ a_j \tau_j^{k+1} \Gamma(k+1) \right] \tag{4}$$

Hence the steady state mean input current $\bar{I}_\infty$ is proportional to the input frequency $\lambda_{in}$.

## 3. Output frequency is a nonlinear function of input current

Suppose the input current is constant, i.e., $I(t) = I_{input}$, and $V(0) = V_{after}$, then integrating both sides of (1) w.r.t. $t$ gives

$$\tau_m \int_{V_{after}}^{V(\Delta t)} dV(t) = \int_0^{\Delta t} \left( -(V(t) - V_{rest}) + RI_{input} \right) dt \tag{5}$$

and whose solution is

$$V(t) = V_{after} + (V_{rest} - V_{after} + RI_{input})(1 - e^{-t/\tau_m}) \tag{6}$$

If a spike occurs at threshold membrane potential $V_{thr}$, then (6) can be solved by setting $V(t) = V_{thr}$ to yield a nonlinear relationship between input current $I_{input}$ and output spike frequency $\lambda_{out}$:

$$\Delta t = -\tau_m \ln\left(1 - \frac{V_{thr} - V_{after}}{V_{rest} - V_{after} + RI_{input}}\right) = \frac{1}{\lambda_{out}} \tag{7}$$

There is a minimum current required to reach threshold:

$$I_{min} = \frac{V_{thr} - V_{rest}}{R} \tag{8}$$

which means that the $I_{input} \to \lambda_{out}$ relationship is not smooth (in addition to being nonlinear).

The spiking neuron model may include a constant absolute refractory period $t_r$ immediately after every output spike, whereby the membrane potential is clamped to $V_{after}$. That is equivalent to a delay of exactly $t_r$ before the beginning of every depolarization phase. Hence the ISI is

$$\Delta t = t_r - \tau_m \ln\left(1 - \frac{V_{thr} - V_{after}}{V_{rest} - V_{after} + RI_{input}}\right) = \frac{1}{\lambda_{out}} \tag{9}$$

## 4. Output spike frequency is a nonlinear function of input spike frequency

Next we apply asymptotic mean current approximation (AMCA) to model postsynaptic spike trains.

The effect of a regular train of input spikes on membrane potential can be approximated by their average current (Fig A3):

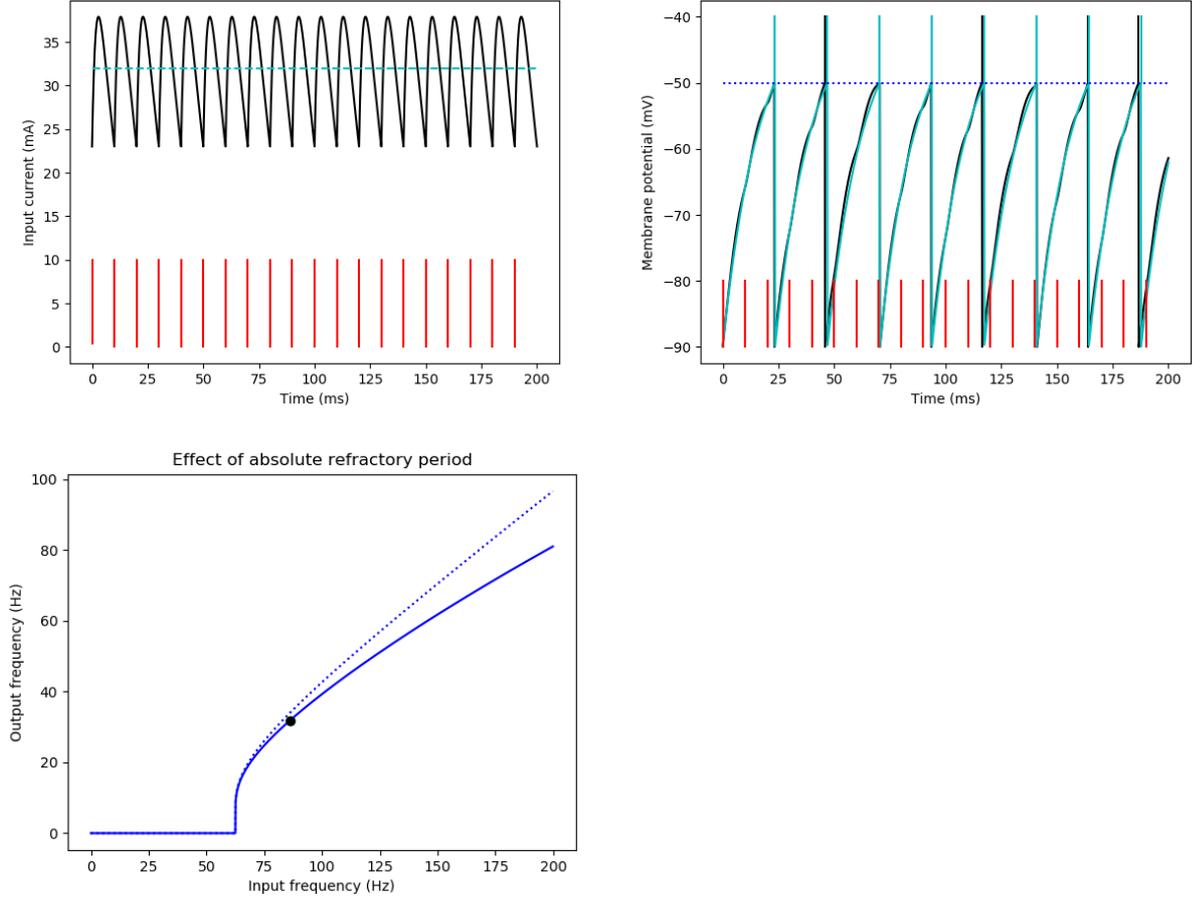

Fig A3: Modelling postsynaptic spike trains using the asymptotic mean current approximation (AMCA). Top left: Regular input spike times (red) lead to periodic synaptic currents (black) and a stable average current (cyan). Top right: Numerically integrated membrane potential (black) is similar to the mean-current-approximation (cyan). Bottom left: Output spike frequency vs input spike frequency, using AMCA, including (solid line) and excluding (dotted line) an absolute refractory period. The result from a numerical simulation with absolute refractory period (black dot) shows the match with AMCA.

We can then combine (4) and (9) to obtain a relationship between the input and output spike frequencies:

$$\lambda_{out} = \begin{cases} \dfrac{1}{t_r - \tau_m \ln\left(1 - \dfrac{V_{thr} - V_{after}}{V_{rest} - V_{after} + R \lambda_{in} \sum_j a_j \tau_j^{k+1} \Gamma(k+1)}\right)} & \text{if } \lambda_{in} > \lambda_{min} \\ 0 & \text{otherwise} \end{cases} \quad (10)$$

where the minimum input frequency is

$$\lambda_{min} = \frac{V_{thr} - V_{rest}}{R \sum_j a_j \tau_j^{k+1} \Gamma(k+1)} \quad (11)$$

## 5. Maximizing input ISI entropy preserves nonlinearity

In the preceding analysis, input spike trains were assumed to be regular and hence have zero interspike interval (ISI) entropy. This means that at subthreshold input frequencies, there is no possibility of output spiking activity. However, if there is ISI variability, then it may be possible that occasional bursts of input activity can still cause spiking in an output neuron. To determine whether this phenomenon can reduce or even eliminate the nonlinearity between input and output frequencies, we repeat the earlier analysis assuming maximal ISI entropy, which is a Poisson spike train.

*Steady state synaptic current mean and variance for a Poisson input train*

Using the synaptic current model of (2), the instantaneous mean current is, i.e.,

$$\begin{aligned} E_T &= E\left(\lim_{\Delta t \to 0} \frac{1}{\Delta t} \int_0^{\Delta t} I_\infty(t)\, dt\right) \\ &= \lim_{\Delta t \to 0} \sum_{j=0}^{\infty} P(spike)\left(\frac{1}{\Delta t}\int_0^{\Delta t} I_1(t+j\Delta t)\, dt\right) \\ &= \lambda_{in} \int_0^{\infty} I_1(t)\, dt \\ &= \lambda_{in} \sum_j \left[a_j\, \tau_j^{k+1}\, \Gamma(k+1)\right] \end{aligned} \qquad (12)$$

where $P(spike)=\lambda_{in}\Delta t$ is the probability of an input spike occurring in any time interval $\Delta t$. Note that $E_T$ denotes the steady state expected current for Poisson input trains, whereas $\overline{I}_\infty$ was used earlier to denote the steady state current averaged over a single interspike interval (ISI) of duration $\Delta t = 1/\lambda_{in}$. While $I_\infty(t)$ is a fixed current profile for regular input trains, it varies for Poisson input trains. Hence $E_T$ and $\overline{I}_\infty$ are derived differently, but yield identical results, i.e., $E_T = \overline{I}_\infty$.

The instantaneous variance of the synaptic current can be derived in a similar way:

$$\begin{aligned} V_T &= V\left(\lim_{\Delta t \to 0} \frac{1}{\Delta t}\int_0^{\Delta t} I_\infty(t)\, dt\right) \\ &= \lambda_{in} \int_0^{\infty} \left[\sum_j a_j t^k e^{-t/\tau_j}\right]^2 dt \\ &= \lambda_{in} \int_0^{\infty} \left(\sum_i \sum_j a_i a_j t^{2k} e^{-t\left(\frac{1}{\tau_i}+\frac{1}{\tau_j}\right)}\right) dt \\ &= \lambda_{in} \left[\sum_i \sum_j \frac{a_i a_j}{-\left(\frac{1}{\tau_i}+\frac{1}{\tau_j}\right)^{2k+1}} \Gamma\!\left(2k+1, t\!\left(\frac{1}{\tau_i}+\frac{1}{\tau_j}\right)\right)\right]_0^{\infty} \\ &= \lambda_{in} \sum_i \sum_j \frac{a_i a_j}{\left(\frac{1}{\tau_i}+\frac{1}{\tau_j}\right)^{2k+1}} \Gamma(2k+1) \end{aligned} \qquad (13)$$

Example 1: if $I_1(t)=I_0\left(e^{-t/\tau_1}-e^{-t/\tau_2}\right)$ then $k=0$, $a_1=I_0$ and $a_2=-I_0$ so (12) reduces to

$$E_T = \lambda_{in} I_0 (\tau_1 - \tau_2) \tag{14}$$

and (13) reduces to

$$V_T = \lambda_{in} I_0^2 \frac{(\tau_1 - \tau_2)^2}{2(\tau_1 + \tau_2)} \tag{15}$$

Example 2: if $I_1(t) = w \frac{t}{\tau} e^{1-t/\tau}$, then $k=1$ and $a = we/\tau$ so (12) reduces to

$$E_T = \lambda_{in} w e \tau \tag{16}$$

and (13) reduces to

$$V_T = \lambda_{in} \frac{w^2 e^2 \tau}{4} \tag{17}$$

*Current and spike frequency distributions*

Since mean currents are strictly non-negative, we approximate the distribution as a truncated Gaussian. If the underlying parameters of the *un*truncated Gaussian are *μ* and *σ*,

$$pdf(\overline{I}; \mu, \sigma) \approx \begin{cases} \frac{1}{\sigma} \frac{\phi\left(\frac{\overline{I}-\mu}{\sigma}\right)}{1 - \Phi\left(\frac{-\mu}{\sigma}\right)} & \overline{I} \geq 0 \\ 0 & otherwise \end{cases} \tag{18}$$

where $\phi(z) = e^{-z^2/2} / \sqrt{2\pi}$ and $\Phi(z) = \frac{1}{\sqrt{2\pi}} \int_{-\infty}^{z} e^{-z^2/2} dz = \frac{1}{2}\left(1 + erf(z/\sqrt{2})\right)$. The average output frequency is

$$\overline{\lambda_{out}} = \int pdf(\overline{I}; \mu, \sigma) \lambda_{out}(\overline{I}) d\overline{I} \tag{19}$$

where

$$\lambda_{out}(\overline{I}) = \frac{1}{t_r - \tau_m \ln\left(1 - \frac{V_{thr} - V_{after}}{V_{rest} - V_{after} + R\overline{I}}\right)} \tag{20}$$

To compute $\overline{\lambda_{out}}$ it is necessary to find the parameters of the truncated Gaussian, $\mu$ and $\sigma$, across all $\lambda_{in}$. Since both $E_T$ and $V_T$ were shown earlier to be directly proportional to $\lambda_{in}$ (see (12) and (13)), it is possible to empirically fit $\mu$ and $\sigma$ to yield the best match for each $E_T$ and $V_T$. A more succinct alternative method is to use the inverse coefficient of variation.

*Using the inverse coefficient of variation*

Let $y=\mu/\sigma$ be the inverse of the coefficient of variation of the untruncated Gaussian. For all $\lambda_{in} > \lambda_c$, $\mu$ and $\sigma$ can be computed from $y$, and hence $\overline{\lambda_{out}}$ can be found according to (19). The minimum $\lambda_c$ can be found as follows.

Using the Poincaré series of erfc() we can write

$$\frac{1}{2} erfc\left(\frac{x}{\sqrt{2}}\right) = \frac{e^{-x^2/2}}{x\sqrt{2\pi}} \sum_{j=0}^{\infty} (-1)^j \frac{(2j-1)!!}{x^{2j}} \tag{21}$$

which, due to higher order terms vanishing, can be used to show that

$$\lim_{y \to -\infty} \left(-y - \frac{1}{y}\right)(1 - \Phi(-y)) = \phi(-y) \tag{22}$$

This implies that $V_T = E_T^2$. The cutoff frequency is thus

$$\lambda_c = \frac{\sum_i \sum_j \frac{a_i a_j}{\left(\frac{1}{\tau_i} + \frac{1}{\tau_j}\right)^{2k+1}} \Gamma(2k+1)}{\sum_i \sum_j a_i a_j \tau_i^{k+1} \tau_j^{k+1} \Gamma^2(k+1)} \tag{23}$$

This is the minimum input frequency which can be computed by using the parameter $y$, corresponding to the limit as $y=\mu/\sigma \to -\infty$.

For $I_1(t) = I_0(e^{-t/\tau_1} - e^{-t/\tau_2})$ (23) reduces to

$$\lambda_c = \frac{1}{2(\tau_1 + \tau_2)} \tag{24}$$

For $I_1(t) = w\frac{t}{\tau}e^{1-t/\tau}$ (23) reduces to

$$\lambda_c = \frac{1}{4\tau} \tag{25}$$

Below $\lambda_c$, $y_{min}$ is a constant computed by fitting $(\mu, \sigma)$ to $(E_T, V_T)$ corresponding to $\lambda_c$.

*Examples of input-output frequency functions*

Example 1) Piecewise construction of input-output frequency function for Poisson input trains using $I_1(t) = I_0(e^{-t/\tau_1} - e^{-t/\tau_2})$ (Fig A4):

1) Low input frequencies: parameterized by input frequency $0 < \lambda_{in} \leq \lambda_c = \frac{1}{2(\tau_1 + \tau_2)}$

$$\begin{aligned}
\kappa_{min} &= \frac{\phi(-\gamma_{min})}{1-\Phi(-\gamma_{min})} \\
V_T &= \lambda_{in} I_0^2 \frac{(\tau_1-\tau_2)^2}{2(\tau_1+\tau_2)} \\
E_T &= \lambda_{in} I_0 (\tau_1-\tau_2) \\
\sigma &= \sqrt{\frac{V_T}{1-\kappa_{min}(\kappa_{min}+\gamma_{min})}} \\
\mu &= E_T - \kappa_{min}\sigma
\end{aligned} \qquad (26)$$

2) High input frequencies: parameterized by $\gamma = \mu/\sigma \geq \gamma_{min}$

$$\begin{aligned}
\kappa &= \frac{\phi(-\gamma)}{1-\Phi(-\gamma)} \\
\lambda_{in} &= \frac{(\kappa+\gamma)^2}{2(1-\kappa(\kappa+\gamma))(\tau_1+\tau_2)} \\
E_T &= \lambda_{in} I_0 (\tau_1-\tau_2) \\
\sigma &= \frac{E_T}{\kappa+\gamma} \\
\mu &= E_T - \kappa\sigma
\end{aligned} \qquad (27)$$

3) Linear interpolation between (1) and (2)

NB: (18) and (19) are used for numerical integration in (1) and (2) to yield mean output frequency.

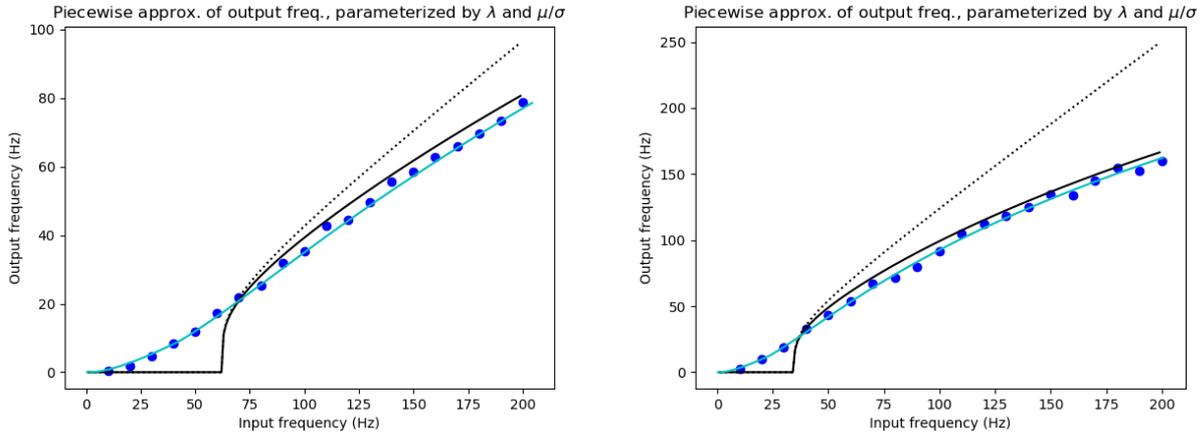

Fig A4: Piecewise construction of input-output frequency function for Poisson input trains. Left: Using $\lambda_c$, the corresponding $E_T$ and $V_T$ are computed, and the corresponding $\mu$ and $\sigma$ are fitted, giving $\gamma_{min}$. Below $\lambda_c$, treat $\gamma_{min}$ and k as constants. Black lines: regular input trains (minimum ISI entropy) either with (solid) or without (dotted) absolute refractory period. Blue dots: SNN simulation results. Cyan line: mathematical model results. Model parameters: $I_0$ = 100pA, $\tau_1$ = 10ms, $\tau_2$ = 2ms, $\tau_m$ = 16ms, $t_r$ = 2ms, R = 400MΩ, $V_{rest}$ = -70mV, $V_{after}$ = -90mV, $V_{thr}$ = -50mV. Right: As for Left but with different model parameters ($I_0$ = 30pA, $\tau_1$ = 20ms, $\tau_2$ = 2ms, $\tau_m$ = 10ms, $t_r$ = 2ms, R = 800MΩ, $V_{rest}$ = -70mV, $V_{after}$ = -90mV, $V_{thr}$ = -55mV).

Example 2) Piecewise construction of input-output frequency function for Poisson input trains using $I_1(t) = w \frac{t}{\tau} e^{1-t/\tau}$:

1) Low input frequencies: parameterized by input frequency $0 < \lambda \leq \lambda_c = \frac{1}{4\tau}$

$$\begin{aligned} \kappa_{min} &= \frac{\phi(-\gamma_{min})}{1 - \Phi(-\gamma_{min})} \\ V_T &= \frac{\lambda_{in} w^2 e^2 \tau}{4} \\ E_T &= \lambda_{in} w e \tau \\ \sigma &= \sqrt{\frac{V_T}{1 - \kappa_{min}(\kappa_{min} + \gamma_{min})}} \\ \mu &= E_T - \kappa_{min} \sigma \end{aligned} \qquad (28)$$

2) High input frequencies: parameterized by $\gamma = \mu/\sigma \geq \gamma_{min}$

$$\begin{aligned} \kappa &= \frac{\phi(-\gamma)}{1 - \Phi(-\gamma)} \\ \lambda_{in} &= \frac{(\kappa + \gamma)^2}{4 \tau (1 - \kappa(\kappa + \gamma))} \\ E_T &= \lambda_{in} w e \tau \\ \sigma &= \frac{E_T}{\kappa + \gamma} \\ \mu &= E_T - \kappa \sigma \end{aligned} \qquad (29)$$

3) Linear interpolation between (1) and (2).

NB: (18) and (19) are used for numerical integration in (1) and (2) to yield mean output frequency.

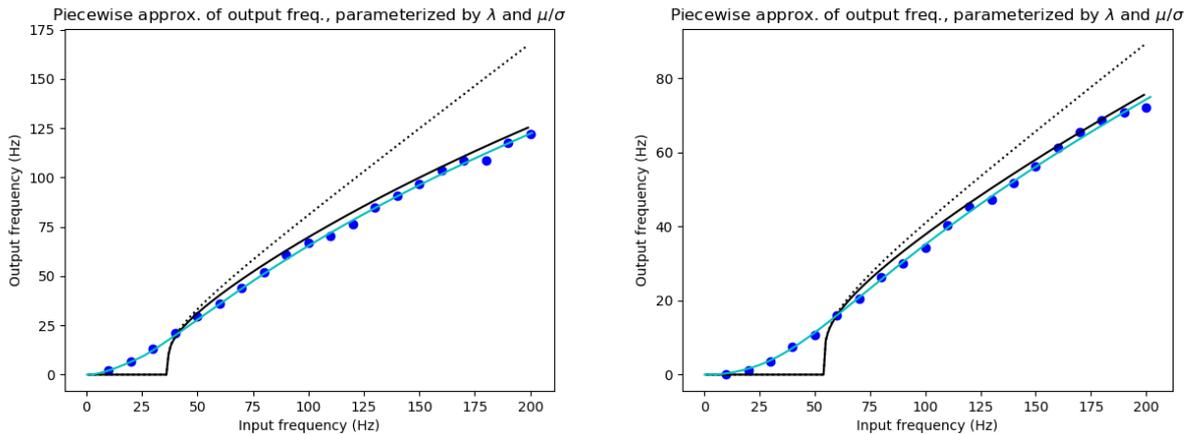

Fig A5: Piecewise construction of input-output frequency function for Poisson input trains. Left: As for Fig A4 using the NEST synaptic model. Model parameters: w = 100pA, $\tau$ = 10ms, $\tau_m$ = 16ms, $t_r$ = 2ms, R = 200MΩ, $V_{rest}$ = -70mV, $V_{after}$ = -90mV, $V_{thr}$ = -50mV. Right: NEST synaptic model with different parameter values (w = 25pA, τ = 12ms, $τ_m$ = 16ms, $t_r$ = 2ms, R = 450MΩ, $V_{rest}$ = -70mV, $V_{after}$ = -90mV, $V_{thr}$ = -50mV).

## 6. Concurrent signal propagation failure and avalanche

At low network gains (Fig 1d), spikes eventually vanish when propagated through multiple layers, ultimately leading to total signal propagation failure. In contrast, when network gains are high, sparse input patterns vanish while dense input patterns avalanche, the latter causing saturated activity. In the case of variable input rates, such as random patterns, it is possible for both failure and avalanche to occur at different times within the same set of neurons (Fig 2a). We now consider the case when network gain is sufficiently high to prevent total signal propagation failure, while mean firing rates are capped to prevent constant saturated activity.

Section 4 (minimum ISI entropy) and Section 5 (maximum ISI entropy) showed that the input-output frequency function is nonlinear. In all cases, $\lambda_{out}:\lambda_{in}$ ratio was lowest for low frequencies. In the case of regular input trains, there was a threshold below which transmission failed immediately. For Poisson input trains, there was a reduction in the frequency, which, if propagated forward, would lead to eventual failure. So assuming that overall firing rates did not increase significantly across layers in a feedforward network (e.g., using ITP), low firing rates tend to produce even lower firing rates in successive layers, until signal propagation failure occurs.

The converse of the above phenomenon is that high frequency inputs are propagated. In the case of multiple input neurons, the highest possible input spike rates occur via temporal synchronization across multiple input neurons because there is no limit from the refractory period. Clearly, even random spike trains will occasionally synchronize by chance. However, if those high frequency input bursts are transmitted preferentially over the lower frequency asynchronous spikes, then multiple feedforward layers in a neural network will tend to propagate temporally correlated activity, even if they are spurious correlations.

Consider a neuron in a feedforward neural network with 784 inputs (MNIST input size). Suppose the inputs are independent Poisson noise, each with an average spike rate of $λ_0$. The total average input spike rate is thus $784λ_0$. In the time period for which $λ_0$ was computed, the probability that any single input neuron reaching some threshold number of spikes, $n_{thr}$, is

$$Pr(n \geq n_{thr} | 1\ input) = \sum_{j=n_{thr}}^{\infty} \frac{e^{-\lambda_0} \lambda_0^j}{j!} \qquad (30)$$

In contrast, the probability that the total input from all 784 input neurons reaches the same threshold is

$$Pr(n \geq n_{thr} | 784\ inputs) = \sum_{j=n_{thr}}^{\infty} \frac{e^{-784\lambda_0} (784\lambda_0)^j}{j!} \qquad (31)$$

Since these are independent random spike trains, then any temporal alignment can be considered a spurious correlation. Threshold is reached far more frequently through spurious correlations between neurons than bursts of activity in any single neuron. For instance, if $n_{thr}=5$ in a given time period at $\lambda_0=0.02$, there are $4.9 \times 10^7$-fold more postsynaptic spikes due to spurious correlations than bursts of activity from the 784 input neurons individually.

The combined effect of transmission failure at low input frequencies and bias towards spurious correlations at high input frequencies leads to the signal propagation failure-and-avalanche phenomenon described in the text and illustrated in Fig 2a.

# Extended Data

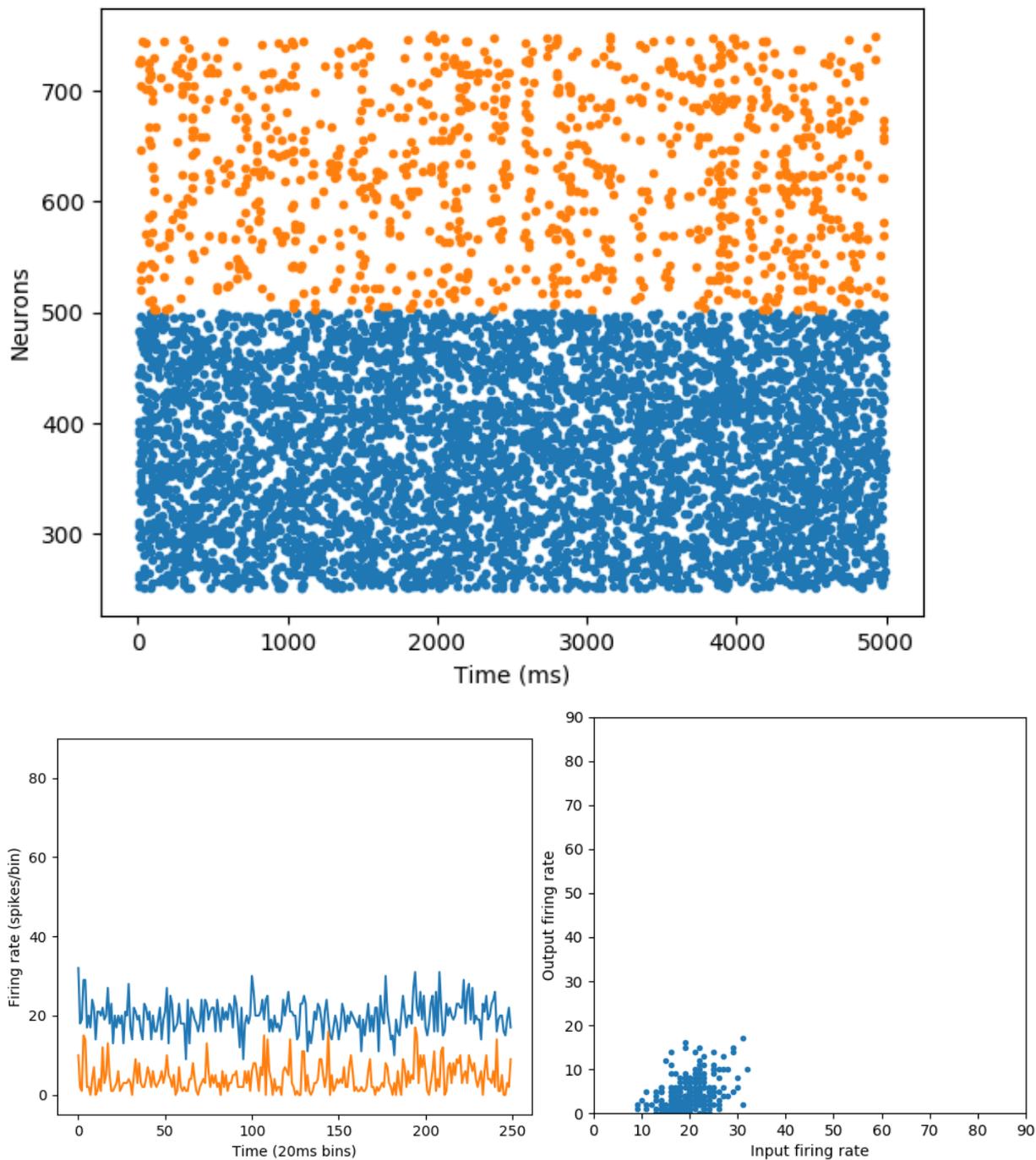

Ext Fig S1: The symptoms of spike propagation failure in a leaky integrate and fire (LIF) neuron model with network gain = 1. The blue population of input neurons (top panel) all fire randomly at a fixed rate (Poisson spike distribution). Each neuron from the orange output population randomly samples from 10% of these input neurons with uniformly distributed connection weights from 0 to a maximum value, where the maximum is tuned to cause *identical absolute changes in firing rate in the outputs as in the inputs*. The output rate tracks changes in the input rate (lower left panel), but with reduced average firing rate. Plotting the input firing rate against the output firing rate (lower right panel) confirms that this effect is analogous to the non-zero threshold of a single neuron.

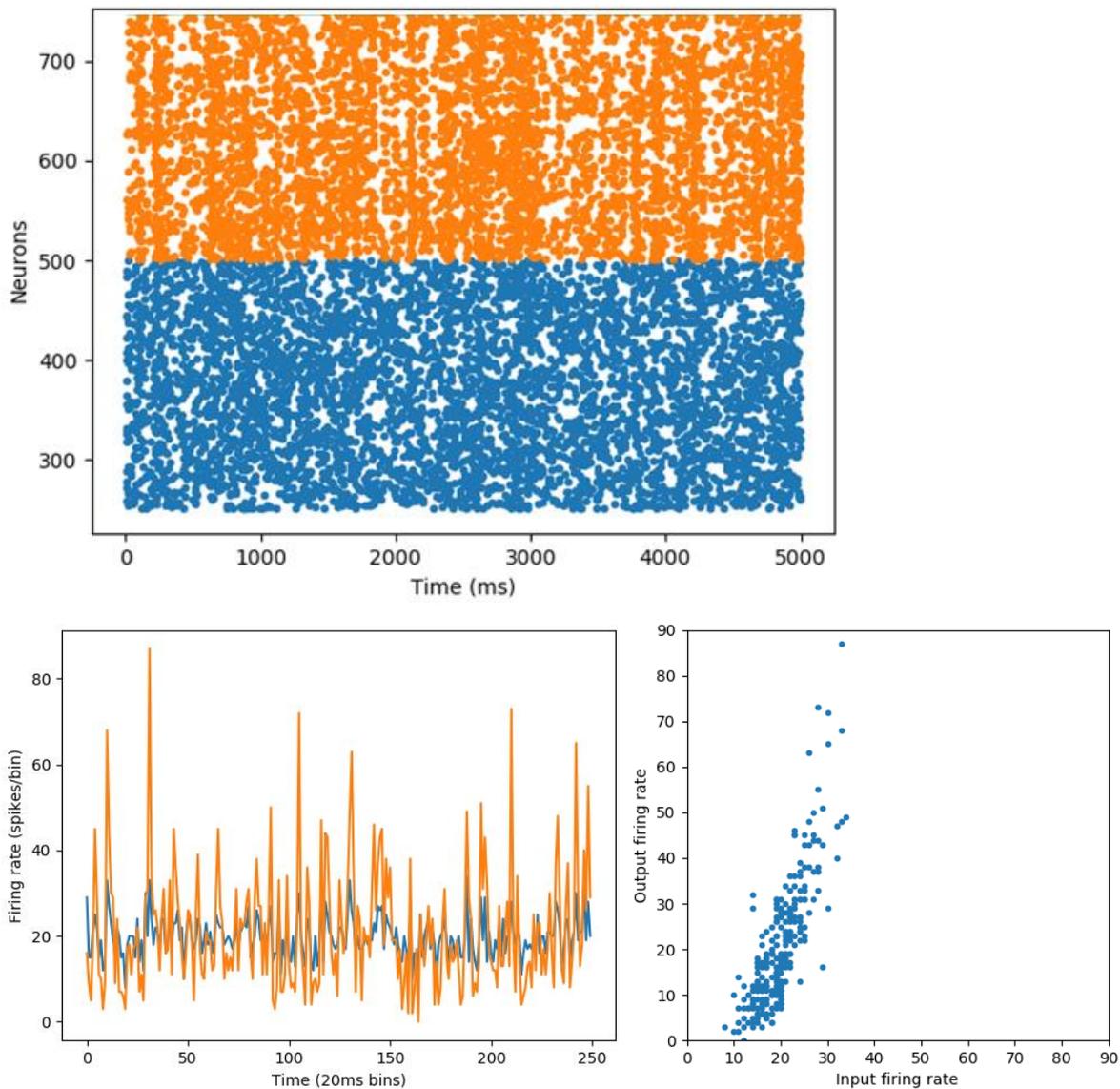

Ext Fig S2: The symptoms of spike propagation failure in a leaky integrate and fire (LIF) neuron model with network gain > 1. The blue population of input neurons all fire randomly at a fixed rate (Poisson spike distribution). Each neuron from the orange output population randomly samples from 10% of these input neurons with uniformly distributed connection weights from 0 to a maximum value, where the maximum is tuned to cause *identical average firing rate in the outputs as the inputs*. Nevertheless, deviations from Poisson firing are clear in the orange output population. In particular, output population bursts occur when the input population rate is randomly higher, and pauses in output occur when the input population rate is randomly lower. This can be seen more clearly in the plot of population firing rates (lower left panel). The output tracks the input rate, but with magnified variability. Plotting the input firing rate against the output firing rate (lower right panel) shows that the slope is greater than 1 and the intercept is not at 0.

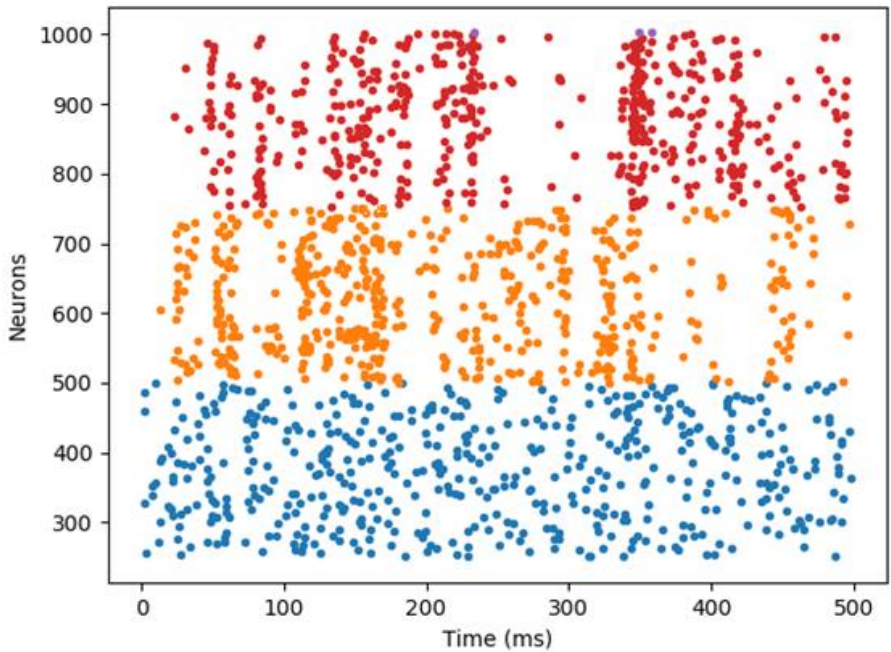

Ext Fig S3: Independent routing of information using the nonlinearity of the spike response. This network contains one input layer (blue) and two output layers (orange and red). Each output layer was connected to a subpopulation of neurons from the input layer. The output layers responded independently, controlled by the number of active neurons to which they were connected in the input layer. In other words, information from the input layer was routed independently based only on the patterns of activity in that layer and the afferent connections. This is a supralinear routing, mimicking winner-take-all mechanisms which are known to be computationally powerful, except here it is accomplished with no additional mechanisms like oscillations or even inhibition. While the depicted routing occurs randomly, it is easy to imagine that the routing could potentially be learned through a mechanism such as reward-modulated STDP, where connections are strengthened when they cause activity patterns that lead to reward.

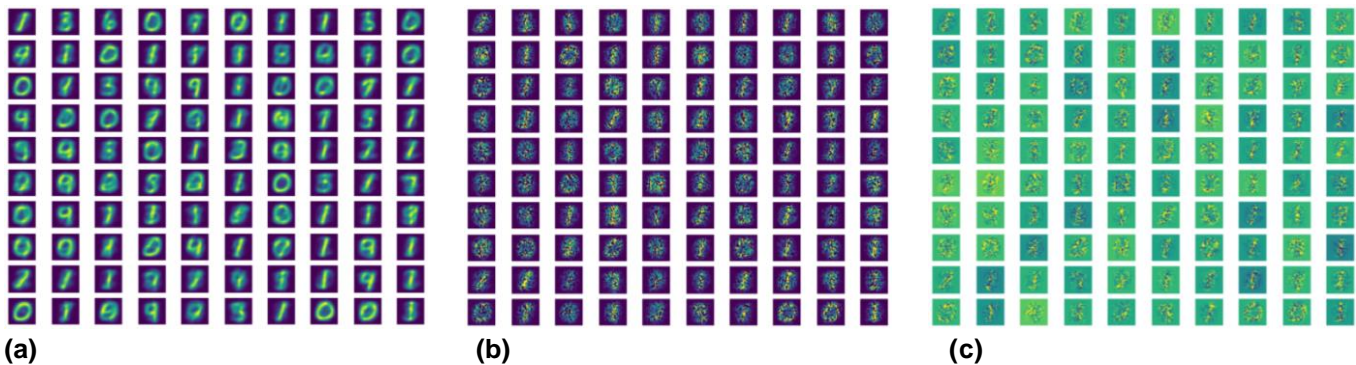

Ext Fig S4: Plastic balancing inhibition learned an approximate negation of the excitatory connections. (a) Excitatory connection weights. (b) Sparse plastic inhibitory connection weights. (c) Sum of (a) and (b) showing that the excitatory digit features were closely balanced by learned inhibition. Inhibitory connection probability was increased to 0.5 for this network to emphasise the balancing effect.

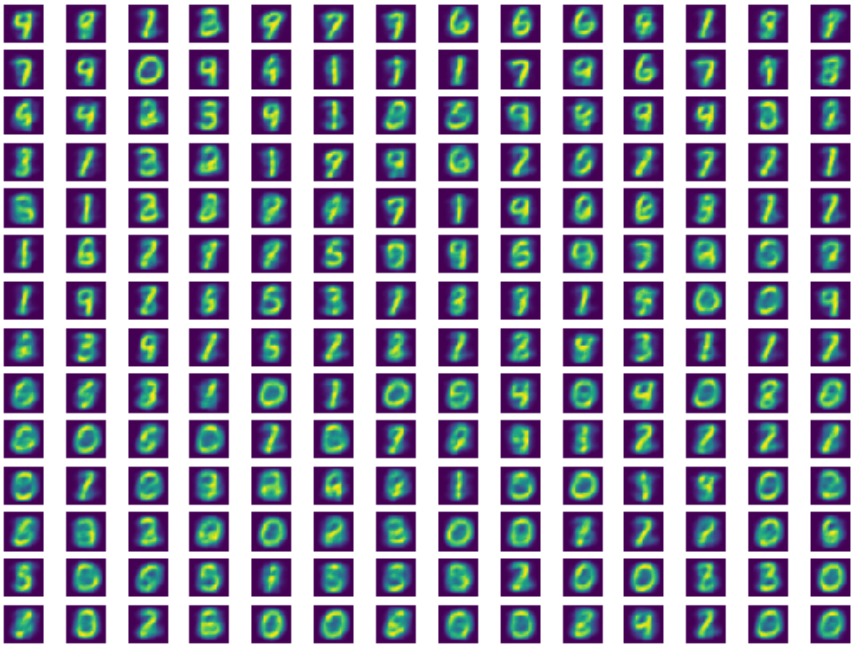

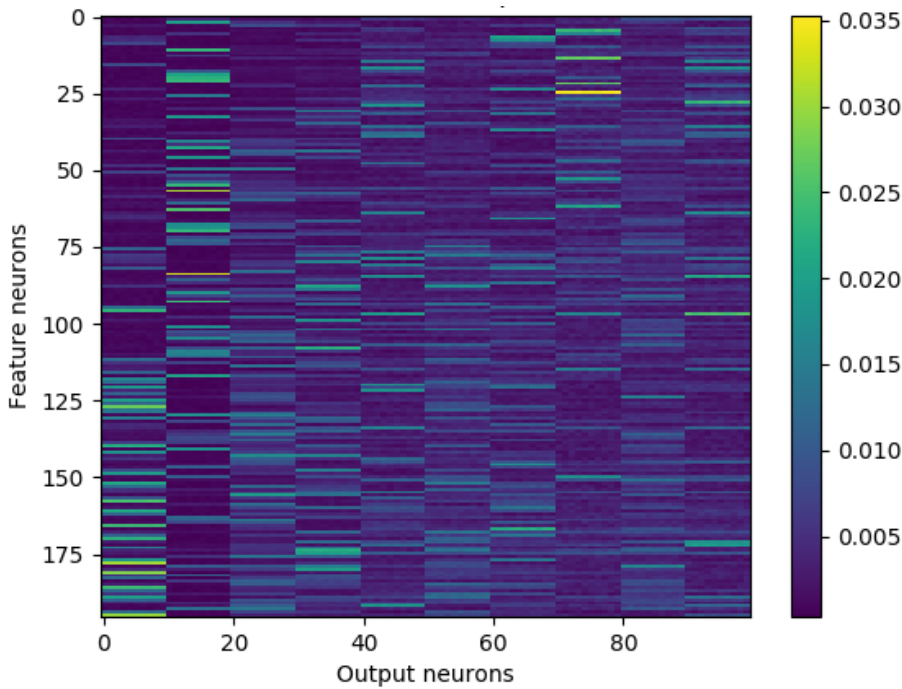

Ext Fig S5: Feature weights in an MNIST network with 196 feature neurons (Top) and 100 output neurons (10 output neurons per digit - Bottom). The feature neurons are sorted in order of ascending firing rate (Top panel – from top left, across then down; Bottom panel – from top to bottom). Output neurons are sorted on output digit (Bottom panel – from left to right; i.e. digit 0 is output neurons 0-9, digit 1 is output neurons 10-19 etc.). As can be seen in both panels, 0's tend to be represented by neurons with mid to high firing rates while 1's are represented by neurons with low to mid firing rates. This occurs because most 0's are similar and therefore appear often, causing neurons with high firing rates to attach to those features. Conversely, 1's are drawn with a wide distribution of orientations, meaning that any given orientation appears only rarely, causing neurons with low firing rates to attach to those features. Most other digits are also biased towards a given firing rate range (e.g. 2's and 3's use mostly higher firing rates; 4's mostly lower; etc.).

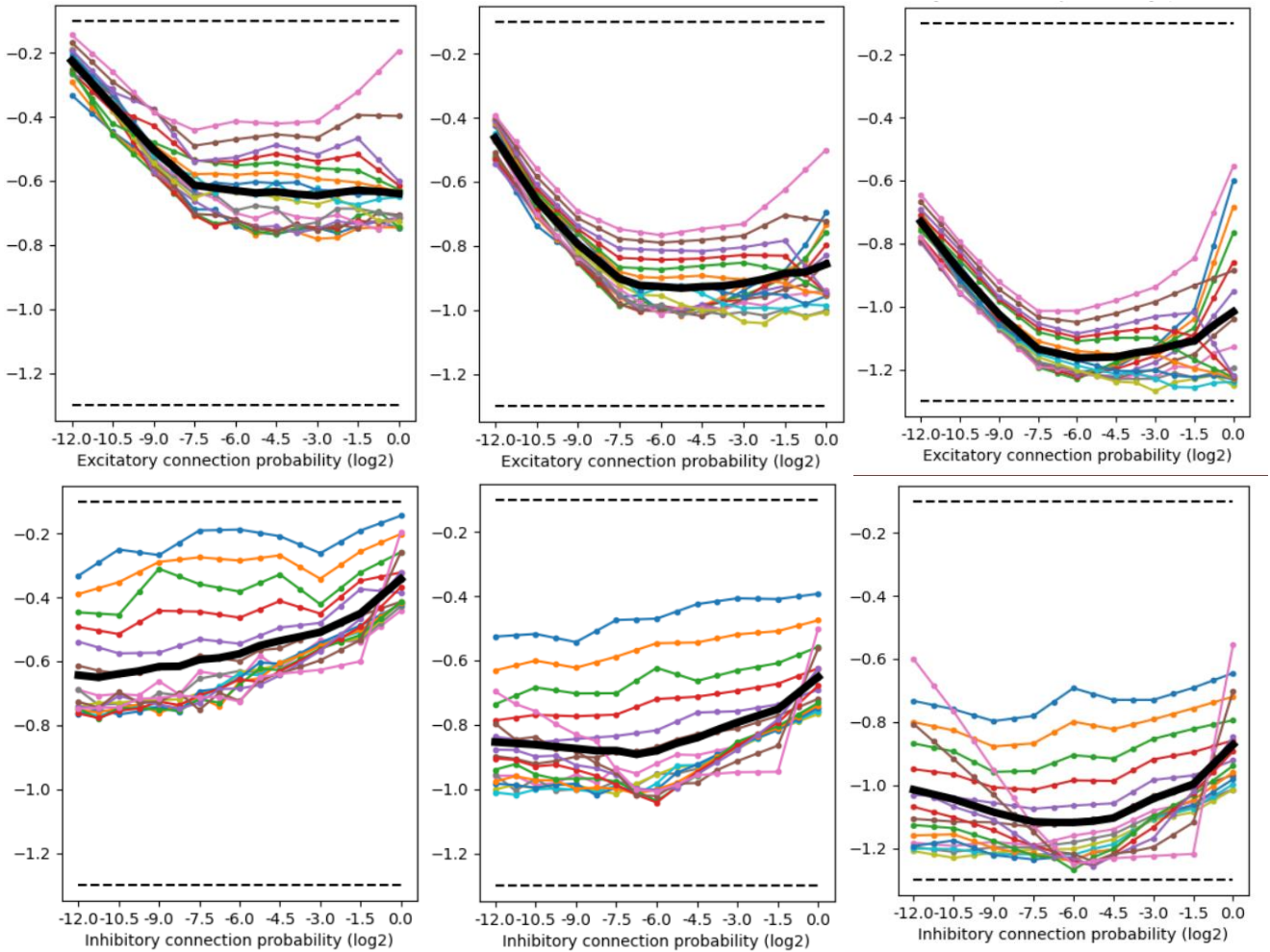

Ext Fig S6: Parameter sweeps across connection probabilities from the MNIST input layer to the feature layer show a sweet spot using sparse connectivity for both excitatory and inhibitory connections. Graphs show log error using a linear decoder on the feature layer (lower values are better). Left panels: 100 feature neurons. Centre panels: 400 feature neurons. Right panels: 1600 feature neurons. Top row: Excitatory connection probability is varied while Inhibitory is held constant (each curve is a constant inhibitory connectivity). Bottom row: Inhibitory connection probability is varied while Excitatory is held constant. Solid black lines: mean performance. The dashed horizontal line at the top of each panel marks chance performance (10%), while the dashed line towards the bottom of each panel marks the 95% correct level. Performance improves with increasing network size. For small networks, maximal excitatory and minimal inhibitory connectivity appear to be optimal. However as network size increases a sweet spot emerges for connection probability around $2^{-6} = 0.016$, meaning that each feature neuron receives on average 12 excitatory and 12 inhibitory connections from the input layer.

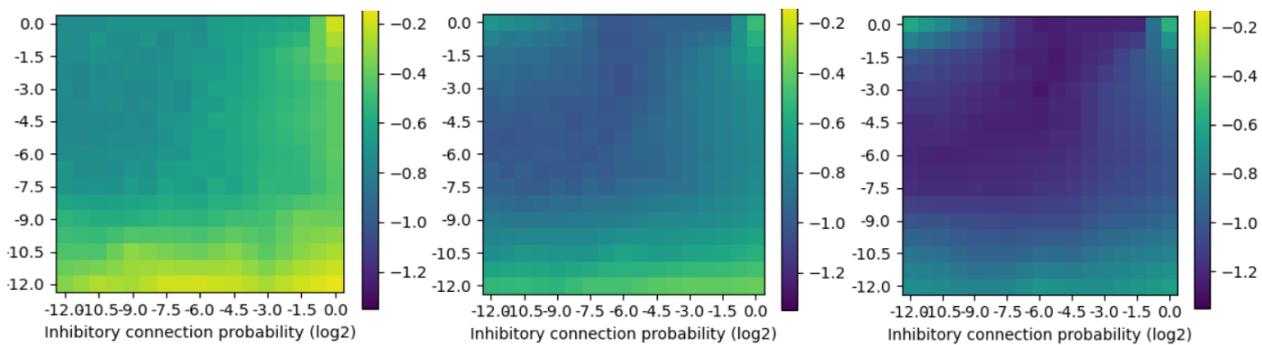

Ext Fig S7: The same information as the previous figure presented as heat maps; excitatory (Y axis) and inhibitory (X axis) connection probabilities vs log error. Yellow = high error; Blue = low error. On these log-log plots, the sweet spot at (-6,-6) that was identified above is clear for the 400 neuron network (centre panel) and the 1600 (right panel). The general conclusions that can be drawn are that:

- An exceptionally broad range of excitatory connectivity from $2^{-6}$ to $2^0$ (0.016 to 1.0) gives good performance as long as inhibitory connectivity is within a certain range.
- There is a sparsity limit below which excitatory connections are too sparse to find all the correlations or hold all the necessary feature information.
- Excitatory and inhibitory connections need to be roughly balanced, but balance that is too good (i.e. full or nearly full excitatory and inhibitory connectivity simultaneously) causes worse performance.

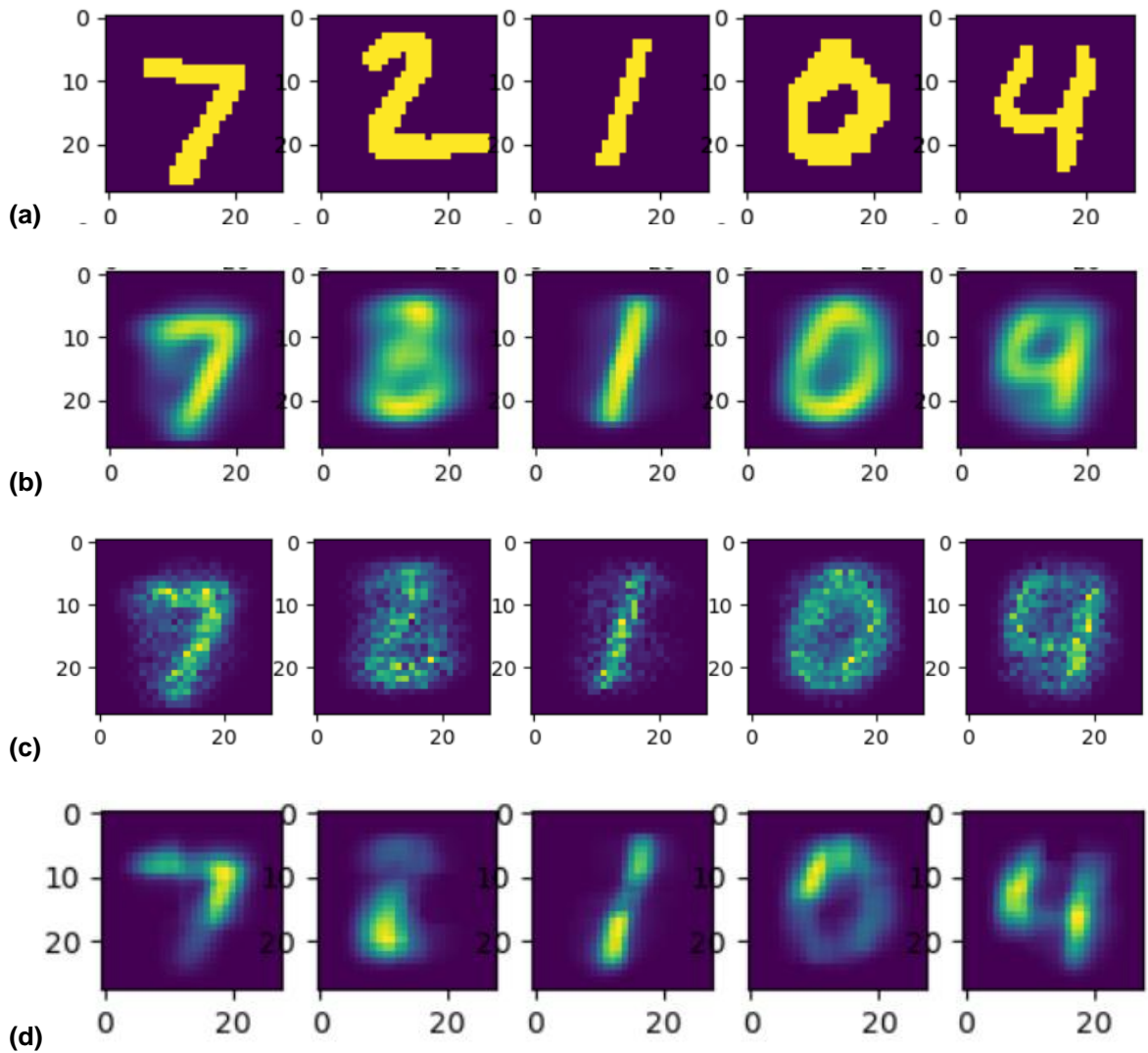

Ext Fig S8: By summing the connection weights of the feature neurons that fire to a given input, it is possible to reconstruct and visualise how the network is perceiving that input. Networks with 3200 feature neurons were trained on MNIST, then the responses of the network to the first five digits in the test set were reconstructed. Since several of these five digits are atypical examples, the reconstruction errors are particularly informative.

(a) First five digits in the MNIST test set (converted to 1 bit per pixel).

(b) Feature layer *fully* connected to the MNIST inputs. Except for the '1' the reconstructed digits are not entirely faithful to the original inputs. The '2' that is reconstructed as a '3' is the most extreme example. This error seems to occur for two reasons. 1. The network has no way of reproducing the long tail because this feature never (or almost never) occurred in the training set. 2. The long tail is also pushing the rest of the digit off-centre and for that reason it seems to be better at activating the features for a '3'. Similarly, the slope of the upper line segment of the '7' is incorrect, the reconstruction of the '0' is too thin, and the reconstruction of the '4' looks somewhat like a '9'. This network correctly classified all digits except the '2' which it misclassified as '3'. Clearly, trying to recognise an entire digit image at once is problematic, since there are too many ways that pixels can be arranged (i.e. a combinatorial explosion) and therefore there are never enough feature neurons to represent all possible combinations. The feature neurons learn 'typical' digits, but they fail to represent all the individual idiosyncrasies that single instances of each digit can have. These idiosyncrasies can cause spurious feature neuron firing that then cause misclassifications.

(c) Feature layer *sparsely* connected to the MNIST inputs. The reconstructed digits are more faithful to the original inputs, but are noisy due to the random sparsity of the connections. The network still failed to represent the long tail of the '2', but in this case it correctly classified all five digits. Using sparse rather than full connectivity resulted in more accurate and less stereotyped reconstructions, since the network was able to activate feature neurons in novel combinations to represent novel inputs.

(d) Each feature neuron *fully connected to a 10x10 region* of the MNIST input space. These spatially restricted receptive fields limited the number of possible pixel combinations within each field to a tractable level. These fields were then repeated across the full extent of the image, somewhat similar to one layer of a convolutional architecture. As for (c) above, this network was able to activate feature neurons in novel combinations to represent novel inputs, but due to the spatially limited extent of each receptive field in this case there was no extraneous noise. The digit reconstructions were the most faithful of all the considered architectures, and the network correctly classified all five digits even though the long tail of the '2' still failed to be reconstructed. This architecture also maximised the classification performance over the entire test set.

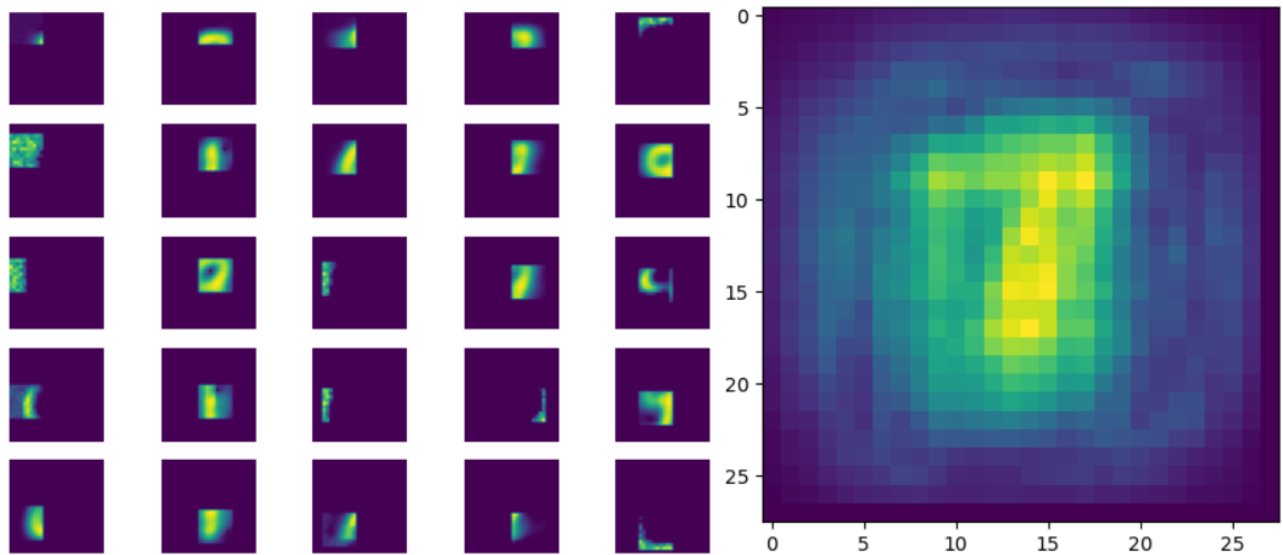

**(a)**

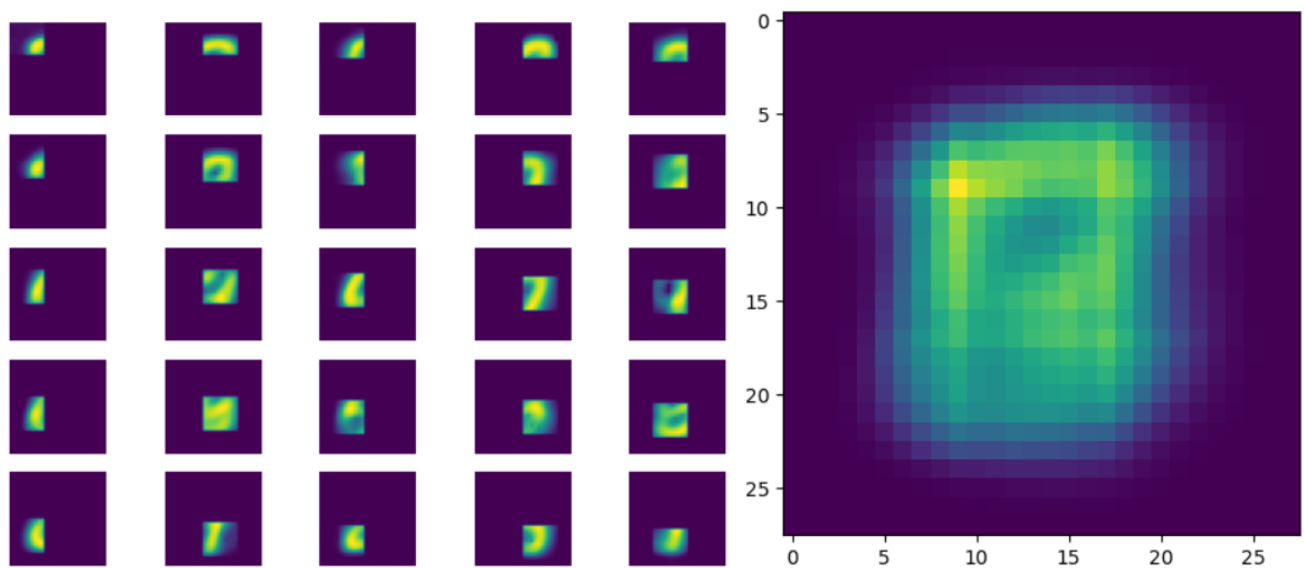

**(b)**

Ext Fig S9: Differences in features with and without constant input to each feature neuron. Note that constant input was not used for the MNIST results in the main manuscript since many MNIST pixels around the edges of the digits are always off, meaning that those feature neurons that developed representations to these blank regions were not helpful for identifying digits. The network outputs easily learned to ignore the unhelpful neurons but overall performance suffered slightly due to there being effectively fewer useful feature neurons. (a) With constant input, some neurons learned to respond to MNIST edge pixels that never actually fired (left panel). These 'features' were wasted since they never helped identify a digit (this wastage is an artefact of the MNIST dataset because it contains input pixels that never fire during training). The sum of all the feature neuron connection weights (right panel) clearly shows the edge pixel responses. (b) Without constant input, each neuron connected only to input pixels that actually fired during training.

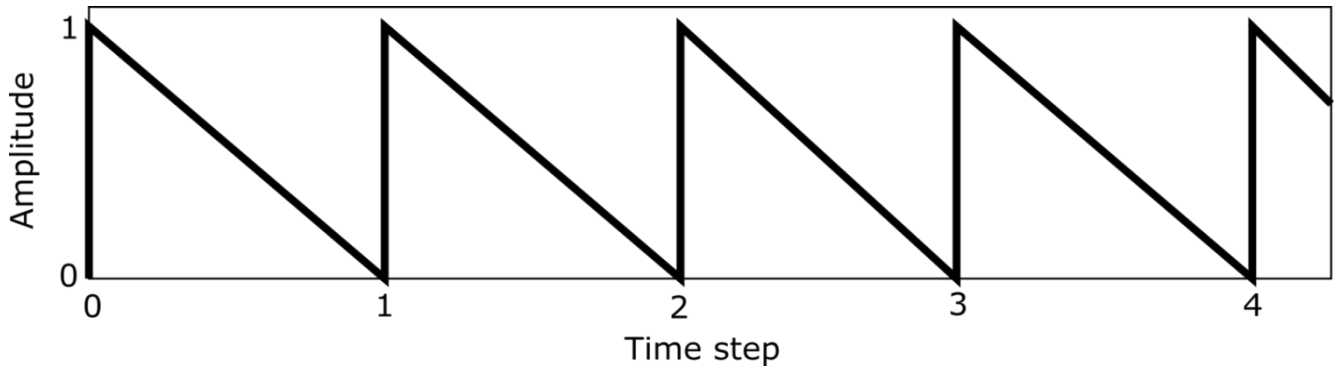

Ext Fig S10: Spikes occur with sub-timestep resolution, where neurons that receive stronger excitatory drive generate larger spikes (that conceptually occur earlier in the timestep) and also propagate stronger drive to downstream neurons.